\documentclass{acm_proc_article-sp}

\usepackage{graphicx}
\usepackage{color}
\usepackage{amsmath}
\usepackage{natbib}
\usepackage{xspace}
\usepackage{xfrac}
\usepackage{tabularx}
\usepackage{multirow}
\usepackage{array}
\usepackage{pbox}

\usepackage[draft]{fixme}
\usepackage{color}

\newcommand{\cWalker}{\textit{c-Walk\-er}\xspace}
\newcolumntype{C}{>{\centering}X}

\begin{document}

\title{Follow, listen, feel and go: \\
alternative guidance systems for \\
a walking assistance device}

\numberofauthors{2} %  in this sample file, there are a *total*
\author{
% 1st. author
\alignauthor
Federico Moro, Antonella De Angeli, Daniele Fontanelli, Roberto Passerone, Luca Rizzon, Stefano Targher, \\Luigi Palopoli\\
       \affaddr{DISI -- University of Trento}\\
       \affaddr{via Sommarive, 9}\\
       \affaddr{I-38056 Povo, Trento, Italy}\\
       \email{federico.moro@unitn.it}
% 2nd. author
% \alignauthor
% Antonella De Angeli\\
%        \affaddr{DISI -- University of Trento}\\
%        \affaddr{via Sommarive, 9}\\
%        \affaddr{I-38123 Povo, Trento, Italy}\\
%        \email{antonella.deangeli@unitn.it}
% % 3rd. author
% \alignauthor Daniele Fontanelli\\
%        \affaddr{DISI -- University of Trento}\\
%        \affaddr{via Sommarive, 9}\\
%        \affaddr{I-38123 Povo, Trento, Italy}\\
%        \email{daniele.fontanelli@unitn.it}
% \and  % use '\and' if you need 'another row' of author names
% % 4th. author
% \alignauthor Roberto Passerone\\
%        \affaddr{DISI -- University of Trento}\\
%        \affaddr{via Sommarive, 9}\\
%        \affaddr{I-38123 Povo, Trento, Italy}\\
%        \email{roberto.passerone@unitn.it}
% 5th. author
\alignauthor Domenico Prattichizzo, and \\Stefano Scheggi\\
       \affaddr{DIISM - University of Siena}\\
       \affaddr{Via Roma 56}\\
       \affaddr{I-53100 Siena, Italy}\\
       % \email{prattichizzo@ing.unisi.it}
% % 6th. author
% \alignauthor Luca Rizzon\\
%        \affaddr{DISI -- University of Trento}\\
%        \affaddr{via Sommarive, 9}\\
%        \affaddr{I-38123 Povo, Trento, Italy}\\
%        \email{luca.rizzon@unitn.it}
% % 7th. author
% \and
% \alignauthor Stefano Scheggi\\
%        \affaddr{DIISM - University of Siena}\\
%        \affaddr{Via Roma 56}\\
%        \affaddr{I-53100 Siena, Italy}\\
%        \email{scheggi@dii.unisi.it}
% % 8th. author
% \alignauthor Stefano Targher\\
%        \affaddr{DISI -- University of Trento}\\
%        \affaddr{via Sommarive, 9}\\
%        \affaddr{I-38123 Povo, Trento, Italy}\\
%        \email{stefano.targher@unitn.it}
% % 9th. author
% \alignauthor Luigi Palopoli\\
%        \affaddr{DISI -- University of Trento}\\
%        \affaddr{via Sommarive, 9}\\
%        \affaddr{I-38123 Povo, Trento, Italy}\\
%        \email{luigi.palopoli@unitn.it}
}
% There's nothing stopping you putting the seventh, eighth, etc.
% author on the opening page (as the 'third row') but we ask,
% for aesthetic reasons that you place these 'additional authors'
% in the \additional authors block, viz.
\date{15 January 2016}
% Just remember to make sure that the TOTAL number of authors
% is the number that will appear on the first page PLUS the
% number that will appear in the \additionalauthors section.
\toappear{}

\maketitle
\begin{abstract}
In this paper, we propose several solutions to guide an older adult
along a safe path using a robotic walking assistant (the \cWalker).
We consider four different possibilities to execute the task.
One of them is mechanical, with the \cWalker playing an active role
in setting the course.
The other ones are based on tactile or acoustic stimuli, and suggest a
direction of motion that the user is supposed to take on her own will.

We describe the technological basis for the hardware components implementing
the different solutions, and show specialised path following algorithms for
each of them.

The paper reports an extensive user validation activity with a quantitative
and qualitative analysis of the different solutions.
In this work, we test our system just with young participants to establish a
safer methodology that will be used in future studies performed with older
adults.

% \keywords{Assistive robotics \and haptic \and passive guidance \and user evaluation}

\end{abstract}

\section{Introduction}
\label{sec:intro}

Ageing is often associated with reduced mobility which is the
consequence of a combination of physical, sensory and cognitive
degrading. Reduced mobility may weaken older adults' confidence
in getting out alone and traveling autonomously in large spaces.
Reduced mobility has several serious consequences including an
increase in the probability of falls and other physical problems,
such as diabetes or articular diseases. Staying at home, people
lose essential opportunities for socialisation and may worsen
the quality of their nutrition. The result is a self-reinforcing
loop that exacerbates the problems of ageing and accelerates
physical and cognitive decline~\cite{Cacioppo2009}.

\begin{figure}[t!]
\begin{center}
\includegraphics[width=0.95\columnwidth]{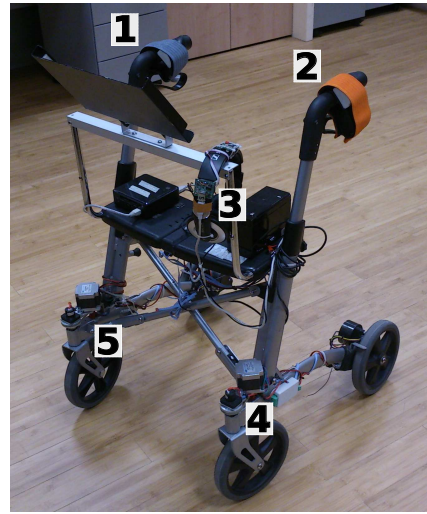}
\caption{The \cWalker with the guidance mechanisms: bracelets, headphone
and steering motors.}
\label{fig:cWalker}
\end{center}
\end{figure}
In the context of different research initiatives (the \textit{DALi}
project\footnote{http://www.ict-dali.eu} and the \textit{ACANTO}
project\footnote{http://www.ict-acanto.eu}) we have developed a
robotic walking assistant, that compensates for sensory and cognitive
impairments and supports the user's navigation across complex
spaces. The device, called \cWalker (Fig.~\ref{fig:cWalker}), is equipped
with different types
of low level sensors (encoders, inertial measurement unit) and
advanced sensors (cameras) that collect information on the device and
its environment. Such measurements are used by the \cWalker to
localise itself and to detect potential risks in the surrounding
environment.  By using this information the \cWalker is able to
produce a motion plan that prevents accidents and drives the user
to her destination with a small effort and satisfying her preferences.
The projects follow an inclusive design approach which requires older users
involvement and participation at appropriate moments in the process once the
evaluation protocols have been validated.

There are different interfaces that the \cWalker can use to guide the user.
Some of them generate acoustic and tactile stimulation
to suggest the correct direction of motion. The user
remains in charge of the last decision on whether to accept or
refuse the suggestions. A different type of mechanisms operates
``actively'' on the walker, by physically changing the direction of
motion.

In this work we describe four different mechanisms for
guidance available in the \cWalker (mechanical, haptic, and two types of
acoustic guidance) showing how they can be
applied in the context of a guidance algorithm and their effect on a student
population. The mechanical guidance is based on the action of
two stepper motors that can change the orientation of the front wheels
forcing a turn in the desired direction.
The haptic guidance is a passive system based on the use of a pair of
bracelets that vibrate in the direction the user is suggested to take.
The same effect can be obtained by administering acoustic signals to
the user through a headphone: a sound on the right side to suggest a
right turn, and a sound on the left side to suggest a left turn.
The acoustic medium has a richer potential. Indeed, by using
appropriate algorithms, it is possible to simulate a sound
in the space
that the user should follow in order to move in the right
direction.

In the paper, we describe the theoretical foundations of the different
mechanisms and algorithms and offer some details and insight on how
they can be integrated in the
\cWalker. In addition, we present the results of two evaluation studies
aimed at providing a protocol for
the evaluation of the performance of the guidance systems and initial
knowledge on user 
behavior and experience. Results suggest that the mechanical performance
is the best and expose
strengths and weaknesses of the other solutions, opening important design
directions for future guidance systems design.

The paper is organised as follows.
In Sec.~\ref{sec:relwork}, we review the most important scientific literature
related to our work.
In Sec.~\ref{sec:architecture}, we describe the hardware and software 
components of the system, while in Sec.~\ref{sec:actuators}, we illustrate the
different guidance mechanisms used in the \cWalker.
In Sec.~\ref{sec:systems}, we describe the guidance algorithms of the
different guidance mechanisms.
We report our testing and validation activities performed with young
participants on all of the systems in Sec.~\ref{sec:experiments}, and finally
we conclude with Sec.~\ref{sec:conclusions}.

\section{Related Work}
\label{sec:relwork}
The robot wheelchair proposed in~\cite{urdiales2011new} offers
guidance assistance in such a way that decisions come from the
contribution of both the user and the machine. The shared control,
instead of a conventional switch from robot to user mode, is a
collaborative control. For each situation, the commands from robot and
user are weighted according to the respective experience and ability
leading to a combined action.  

Other projects make use of walkers to provide the user with services
such as physical support and obstacle
avoidance. In~\cite{chuy2006new}, the walker can work in
\textit{manual mode} where the control of the robot is left to the
user and only voice messages are used to provide instructions. A
shared control operates in \textit{automatic mode} when obstacle
avoidance is needed and user intention is overridden acting on the
front wheels.

The two projects just mentioned can be considered as ``active''
guidance systems, meaning that the system actively operates to steer
the user toward the desired direction. The \cWalker's mechanical
guidance considered in this paper falls in the same category.  Another
point of commonality is in the strict cooperation between the system
that generates suggestions and the user that has to implement these
decisions.
In the \cWalker, the user comfort has the same
importance as the accuracy and efficiency of the guidance solution.
In fact, not only does the user provides motive
power but she can also decide to override the
system's decisions forcing her way out of the suggested path.

Key to any guidance system of this kind is the ability to detect and
possibly anticipate the user's intent.  A valuable help in this
direction can be offered by the use of force
sensors. \cite{wasson2003user} use force sensors to modify the
orientation of the front wheels of a walker in case of concerns about
comfort and safety of the user motion. In~\cite{chuy2007control} an
omnidirectional mobile base makes possible to change the centre of
rotation to accommodate user intended motion.  Contrary to these
projects, the \cWalker is intended as a low cost system, for which
expensive force sensors are not affordable. The user intent is
inferred indirectly by observing gait, and by estimating her emotional
state.

More similar to our ideas, is the \textit{JAIST active robotic walker}
(JaRoW), proposed by~\cite{lee2010design}, which uses infrared sensors
to detect lower limb movement of the user and adapt direction and
velocity to her behaviour.

A possible idea to reduce intrusivenenss is to use passive devices,
where suggestions on the direction of motion take the form of visual,
auditory or tactile stimuli, and the user remains totally in charge of
the final decision. Haptic interfaces can be used as a practical
method to implement this idea. Successful stories on the use of haptic
interface can be found in the area of teleoperation of vehicles for
surveillance or exploration of remote or dangerous areas. For this
type of applications, haptic interfaces are used to provide feedback
on sense of motion and the feeling of presence, as
in~\cite{Arias2010}. Similar requirements can be found in rescue
activities, where the robot helps the user to move in environments
where visual feedback are no longer available
(\cite{ghosh2014following}). In the latter application the robot
provides information on its position and direction to the user in
order to help the user follow the robot.
Guidance assistance can be provided by giving feedback on the matching
between the trajectory followed by the user and the planned
trajectory.  In~\cite{ScMoPr-TH14}, a bracelet
provides a warning signal when a large deviation with respect to the
planned trajectory is detected.  In~\cite{ErVeJaDo2005} a belt with
eight tactors is used to provide direction information to the user in
order to complete a way-point navigation plan.  As shown below, haptic
bracelets are one possible guidance method offered by the \cWalker.

Another ``passive'' guidance system is based on acoustic signalling
and is well--suited to users with partial or total visual
disabilities.  Acoustic guidance can be achieved by synthesising a
sound from a virtual point in the direction the user should move
toward. A key element of this method is the ability to efficiently and
accurately render sound signals from a specified point. The main method
to achieve this is based on the Head Related Transfer Function (HRTF)
which changes and needs to be determined for each individual, as
explained in~\cite{blauert}. It represents the ears response for a
given direction of the incoming sound. Other approaches are based on
the modelling of the sound propagation. In the modelling process,
attenuation of the sound is taken into account using the
Interaural Level Difference (ILD) which considers the presence of the
listener head. Instead, Interaural Time Difference (ITD) considers the
distance between ears and sound source. These filtering
processes are computationally demanding. The acoustic guidance
mechanism implemented in the \cWalker is based on the adoption of
lightweight algorithms amenable to an embedded implementation and
detailed in~\cite{rizzon2013embedded,rizzon2014spatial}.

% describe the software and hardware architecture of the device and how they
% provide the implementation of the presented guidance systems
\section{System architecture}
\label{sec:architecture}

The \cWalker hardware and software architecture has been designed
for an easy integration of heterogeneous components (possibly developed
by different teams).

\begin{figure}

\includegraphics[width=\columnwidth]{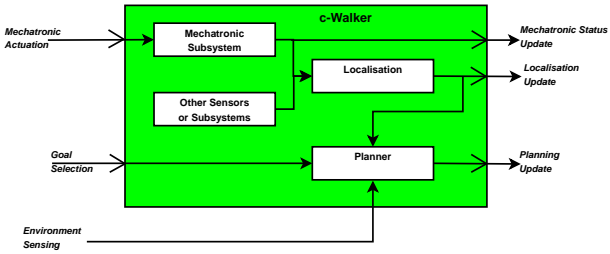}
\caption{The block-scheme of the c-Walker architecture with its core
components: mechatronic subsystem, localisation subsystem and planner.}
\label{fig:cwalker-arch}
\end{figure}

The core modules of the architecture are shown in
Figure~\ref{fig:cwalker-arch}.  The {\tt Planner}\/ decides the plan
to be followed based on: 1. the requests and the preferences of the
user, 2. the map of the environment, 3. the presence of obstacles and
crowded areas along the way. 
While the \cWalker is moving, it collects information from the environment 
and the planned path can be updated to avoid obstacles or safety
risks~\cite{colombo2013motion}.

The Localisation module integrates information from several sources
(encoders, Inertial platform, cameras, RFID reader) to produce an
updated information on the estimated position of the vehicle in the
environment with a few centimetres position~\cite{Naz14}.  A
mechatronic subsystem encapsulates all the modules that are used to
read and process sensor data from the encoders and from the inertial
sensors. Additionally, the mechatronic module contains all the logic required
to send command to the actuators (e.g., the motors on the caster wheels).
The mechatronic system is reached through a CAN bus.

These core modules can be interconnected with other modules to implement the
different guidance solutions discussed above, as shown in
Fig.~\ref{fig:cwalker-arch}.

\begin{figure*}
\centering
\includegraphics[width=0.6\textwidth]{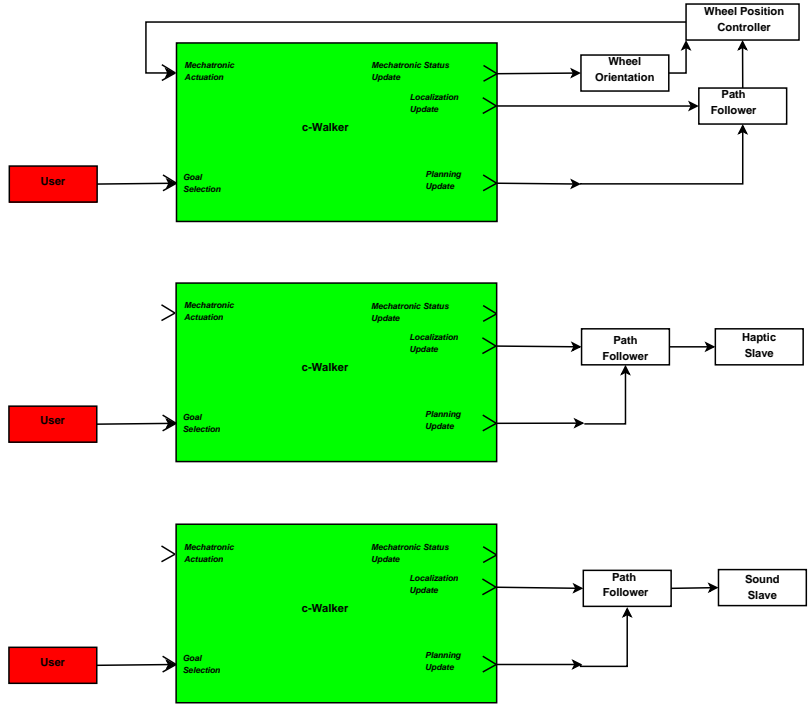}
\caption{The block-scheme of the three guidance systems. The three solutions
interact with the c-Walker by means of the same software interface, but then
are diversified in the way they perform the actuation.}
\label{fig:guidance-arch}
\end{figure*}

The different components are interconnected using a
publisher/subscriber middleware, whereby a component can publish
messages that are broadcast through all the diffferent level of
networking (CAN bus for mechatronic components, ethernet for high
level sensors and computing nodes) in the \cWalker.  This is a key
enabler for the adoption of a truly component--based paradigm, in
which the different guidance systems can be obtained by simply tuning
on some of the modules and allowing them to publish messages or
subscribe to messages.
Three different configurations are schematically shown in
Fig.~\ref{fig:guidance-arch}.

Fig.~\ref{fig:guidance-arch} shows how the three guidance systems that we
present in this work interact with \cWalker.
The scheme on the top of the figure refers to mechanical guidance. The
Planner periodically publishes updated plans (i.e., the coordinates of
the next points to reach). This information subscribed to by a path
follower that implements the algorithm presented in
Section~\ref{subsec:mech}. This component decides a direction for the
wheel that is transmitted to a {\tt Wheel Position Controller} using
the Publish/Subscribe middleware. This component also receives
real--time information on the current orientation of the wheel and
decides the actuation to set the direction to the desired position.
The schemes on the middle and on the bottom apply, respectively, to
haptic and acoustic (and binaural) guidance.
In this case the {\tt Path Follower Components} implements the algorithms 
discussed in Section~\ref{sec:haptic_sound} and transmits its input either the
{\tt Haptic Slave} (see Section~\ref{subsec:bracelets}) or the the {\tt Audio 
Slave} (see Section~\ref{subsec:audio_interface}).

% contains description of the actuators
\section{Guidance mechanisms}

\label{sec:actuators}
In this section, we describe the three main mechanisms that can be used
as ``actuators'' to suggest or to force changes in the direction of motion.

\subsection{Bracelets}
\label{subsec:bracelets}

Haptic guidance is implemented through a tactile stimulation that
takes the form of a vibration. A device able to transmit haptic
signals through vibrations is said ``vibrotactile''.

Vibration is best transmitted on hairy skin because of skin thickness and
nerve depth, and it is best detected in bony areas.
Wrists and spine are generally the preferred choice for detecting vibrations,
with arms immediately following.
Our application is particularly challenging for two reasons:
I.~the interface is designed to be used by older adults,
II.~the signal is transmitted while the user moves.
Movement is known to affect adversely the detection rate and the response time
of lower body sites (\cite{Karuei-2011}).
As regards the perception of tactile stimuli by older adults,
\cite{gescheider94} present studies on the effects of aging in the sense of
touch, which revealed that detection thresholds for several vibration
intensities are higher in older subjects in the age class $65+$. 

Bearing in mind these facts, we designed a wearable haptic bracelet in
which two cylindrical vibro--motors generate vibratory signals to warn
the user (Fig.~\ref{fig:cWalker}).
The subject wears one vibrotactile bracelet on each arm in order to maximize
the stimuli separation while keeping the discrimination process as intuitive as
possible.
In particular, vibration of the left wristband suggests the participant to
turn left, and vice versa.

On each bracelet the distance between the two motors is about $80$~mm.
In two-point discrimination, the minimal distance between two stimuli to be
differentiated is about $35$~mm on the forearms and there is no evidence for
differences among the left and right sides of the body, according
to~\cite{Weinstein68}.
In order to reduce the \emph{aftereffect} problem typical of continuous
stimuli and to preserve users' ability to localize vibration, we selected a
pulsed vibrational signal with frequency $280$~Hz and amplitude of $0.6$~g,
instead of a continuous one.
In particular, when a bracelet is engaged its two vibrating motors
alternatively vibrates for $0.2$~s. 
The choice of using two vibrating motors instead of one was the effect of a
pilot study in which a group of older adults tested both options and declared
their preference for the choice of two motors.
The choice of frequency and amplitude of the vibrations was another outcome of
this study (see~\cite{ScAgPr-IMEKO14}).

From the technological point of view, two \textit{Precision Microdrives}
$303-100$ \textit{Pico Vibe} $3.2$~mm vibration motors were placed into two
fabric pockets on the external surface of the bracelet (the width of the
wristband is about $60$~mm), with shafts aligned with the elbow bone.
The motors have a vibration frequency range of $100$~Hz - $280$~Hz (the
maximal sensitivity is achieved around $200$~Hz - $300$~Hz), lag time of $21$~ms, rise time of $32$~ms and stop time of $35$~ms.

\subsection{Audio interface}
\label{subsec:audio_interface}

The acoustic interface communicates to the user the direction to take 
by transmitting synthetic signals through a headphone (Fig.~\ref{fig:cWalker}).
For instance, when the system aims to suggest a left turn to the user, it
reproduces a sound that is perceived by the user as coming from a point on her
left aligned with the direction she is supposed to take.
This is possible thanks to the application of the binaural theory.

The software module that generates this sound is called {\tt Audio
Slave} and it receives from a master the spatial coordinates
$(S_x,~S_y)$ of the point that is required to be the source of the
sound.  The audio slave converts the cartesian coordinates into a pair
$(r, \theta)$ of relative polar coordinates, in which $r$ represent
the distance between the virtual sound source and the centre of the
listener's head, and $\theta$ represent the azimuthal angle. The pair
$(r, \theta)$ univocally identifies the position of the sound source
on the horizontal plane.  $\theta$ takes on the value $0$ when the
source is in front of the user, positive angles identify positions on
the right hand side, and negative values of $\theta$ identify
positions on the left of the listener.  The guidance signal is a white
noise with duration $50$~ms, which is repeated every $150$~ms.  The
binaural processing algorithm has been used to implement two different
versions of the guidance interface:

\begin{itemize}

\item Left/Right Guidance;

\item Binaural Guidance.

\end{itemize}

Using the Left/Right Guidance Interface, the system reproduces only
virtual sources placed at $\theta=90^\circ$ or at $\theta=-90^\circ$
to suggest a right turn or a left turn, in the same way as the haptic
interface.  With the Binaural Guidance Interface, a virtual sound
source is allowed to be in any position. The resulting suggestion is
not merely for a turn, but it specifies a finer grained information on
the exact direction.  In this case, to ensure the correct displacement
of the virtual sound relative to the user head orientation, an
Inertial Measurement Unit (IMU) monitors the listener's head position
with respect to the \cWalker.
% The IMU sends to the audio software application new position every $250$~ms.
%The angular position of the sound source is discretised in seven
%possible locations on the front hemisphere, while the relative angular
%position corrected by the IMU can take any real value depending on
%the listener's head movements.  Spatial location on the back are
%synthesized as $\theta=\pm90^\circ$ to tell the user which direction
%to take to reach the path and avoid front/back confusion that commonly
%affects binaural sound recognition~\cite{Begault01JAES}. 

Both the interface implementations are based on the same sound
rendering engine which is based on the physics of sound waves. Each of
the sound samples is delayed, and attenuated according to the
principles of sound wave propagation. The binaural effect is obtained
by proper filters that reproduce the presence of the listener's head
and consider the ears displacement.  However, the guidance interface
is meant to generate recommendations on the direction to follow;
therefore, stimuli have been processed without reverberation. As a
consequence, users will perceive the sounds as intracranial, since the
absence of reverb makes it difficult to externalize virtual sound
stimuli.
%In both the audio interface implementations, the sound signals at the
%two channels are delayed, and attenuated according to the relative
%position of the virtual sound source and the listener's ears in the
%virtual space.

\subsection{Mechanical Steering}
\label{subsec:steering}

The mechanical system based on steering uses the front caster wheels to
suggest the user which direction to follow. The positioning of the
wheels causes the \cWalker to perform a smooth turn manoeuvre without
any particular intervention from the user and therefore is considered
as an active guidance. That is, the user provides only the necessary
energy to push the vehicle forward.  Other active approaches
exploiting the braking system acting on the back wheels of the walker
have been presented recently in the literature, among
which~\cite{FontanelliGPP13,saida2011development}.  However, the front
wheels steering approach is more robust and less demanding, in terms
of processing power and sensor measures requests and was then adopted
for this paper.

More in depth, the \cWalker is endowed with two caster wheels in front
of the device, which are connected to a swivel that enables them to
move freely around their axis. Taking advantage of this feature, we
applied steppers motors to the joints to change the
direction of the wheels by a specified amount.  The presence of non-idealities
(e.g., friction and slippage of the gears) can possibly introduce a deviation
between the desired rotation angle and the actual one.
Therefore, we need a position control scheme
operating with real--time measurements of the current angular position
of the wheels. Such measurements are collected by an encoder that is
mounted on the same joint as the stepper motor.  The connection
between wheel and motor is through a gear system such that a complete
turn of the wheel is associated with 4 turns of the motor. Every
complete turn of the motor is $400$ steps.  The stepper motor and the
encoder are controlled by a small computing node that is interfaced to
the rest of the system through a CAN bus.  The motor, together with
the absolute encoder and the relative CAN bus node, is visible in
Fig.~\ref{fig:cWalker}.

With a fixed periodicity, the node samples the encoder and broadcasts
the sensor reading through the bus. The node can also receive a CAN
message coming from other computing devices that requires a rotation
of the wheels specifying the number of degree of rotation and the
angular velocity (deg/s).  The values are automatically converted in
steps and used in a PID control loop that moves the wheel to the
specified angular position.

% contain the description of the three systems
\section{Guidance algorithms}
\label{sec:systems}

The guidance algorithms relies on an accurate estimate of the position
of the \cWalker with respect to the planned path. Since the latter is
generated internally by a module of the \cWalker (see
~\cite{ColomboFLPS13}), only the knowledge on the position
$Q = [x\ y]^T$ and of the orientation $\theta$ expressed in some known
reference frame is needed.  This problem, known in the literature as
{\em localisation problem}, is solved in the \cWalker using the
solutions proposed in~\cite{NazemzadehFM13,NazemzadehFMP14}.

\begin{figure}[t]
\centering
\begin{tabular}{c|c}
  \includegraphics[width=0.45\columnwidth]{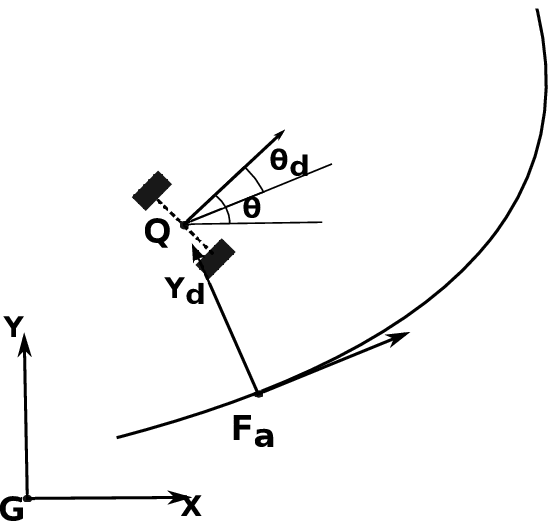} &
  \includegraphics[width=0.45\columnwidth]{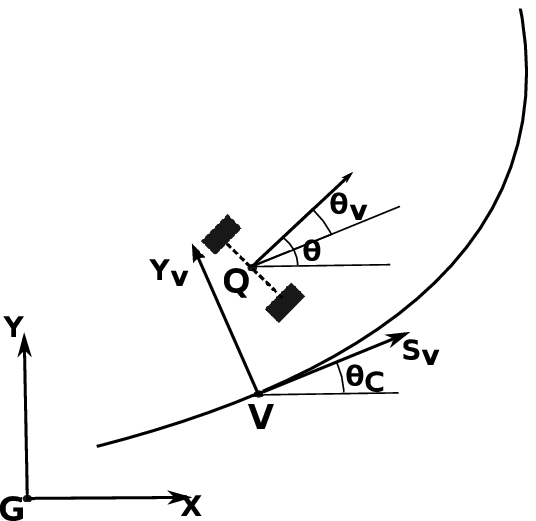} \\
  (a) &
  (b) \\
\end{tabular}
\caption{Graphic representation of the localisation of the \cWalker
with respect to the path:
(a) Frenet-Serret reference,
(b) Virtual vehicle reference. In the Frenet-Serret reference frame, the
vehicle always lies on the $y_d$ axis.}
\label{fig:loc_frame}
\end{figure}

With this information it is possible to determine the Frenet-Serret point
$F_a$, that is a point on the path representing the intersection
between the projection of the vehicle and a segment that is
perpendicular to it and tangent to the path, as in
Fig.~\ref{fig:loc_frame}~(a).
%-%
%-%
We define as $y_d$ and $\theta_d$ respectively the distance along the
projection of the vehicle to $F_a$ and the difference between the
orientation of the \cWalker and the orientation of the tangent to the
path in the projection point. All the proposed guidance algorithms use
this information to compute the specific ``actuation''.

We observe that the objective of the guidance algorithms
is not the perfect path following of the planned trajectory. In fact,
such an objective would be very restrictive for the user and perceived
as too authoritative and intrusive.  In order to give the user the
feeling of being in control of the platform, she is allowed an error (in both,
position and orientation) throughout the execution of the path that
is kept lower than a desired performance threshold.  Therefore,
the path can be considered as the centre line of a virtual corridor in which
the user can move freely.

\subsection{Haptic and Acoustic algorithms}
\label{sec:haptic_sound}

The haptic and acoustic guidance algorithms generate a quantised
control action, which can be described with an alphabet of three
control symbols: a) turn right; b) turn left; c) go straight. This is
a good compromise between accuracy and cognitive load for the
interpretation of signals.

The symbol to be suggested to the user is determined by the
desired turning towards the path.  A straightforward way to compute
such a quantity is to determine the angular velocity an autonomous
robotic unicycle--like vehicle would follow in order to solve the path
following problem.  To this end, we have designed a very simple
control Lyapunov function which ensures a controlled solution to the
path following in the case of straight lines acting only on the
vehicle angular velocity and irrespective of the forward velocity of
the vehicle.  Such a controller works also for
curved paths if we are only interested on the sign of the desired
angular velocity.

To see this, consider the kinematic model of the unicycle
(which is an accurate kinematic model of the \cWalker)
\begin{equation}
\label{eq:UniModel}
\begin{aligned}
\dot x_d & = \cos(\theta_d) v, \\
\dot y_d & = \sin(\theta_d) v, \\
\dot \theta_d & = \omega, \\
\end{aligned}
\end{equation}
where $v \neq 0$ is the forward velocity and $\omega$ its angular
velocity.  $y_d$ and $\theta_d$ are the quantities defined in the
previous section, while $x_d$ is the longitudinal coordinate of the
vehicle that, in the Frenet-Serret reference frame is identically zero
by definition.  It has to be noted that $(x_d,\, y_d)$ are then the
cartesian coordinates, in the Frenet-Serret reference frame, of the
midpoint of the rear wheels axle.  In light of 
model~\eqref{eq:UniModel} and remembering that $x_d$ does not play any
role for path following, we can set up the following control
Lyapunov function
\begin{equation}
\label{eq:lyapunov1}
V_1 = \frac{k_y y_{d}^2 + k_{\theta} \theta_{d}^2}{2} ,
\end{equation}
which is positive definite in the space of interest, i.e., $(y_d,\,
\theta_d)$, and has as time derivative
\begin{equation}
\label{eq:lyapunov1_der}
\dot{V_1} = k_y y_d \sin(\theta) v + k_{\theta} \theta \omega ,
\end{equation}
where $k_y > 0$ and $k_{\theta} > 0$ are tuning constants. Imposing
$\omega$ equals to the following desired angular velocity
\begin{equation}
\label{eq:Omega}
\omega_d =  -q_{\theta} \theta_d - \frac{k_y}{k_{\theta}} y_d \frac{\sin(\theta_d)}{\theta_d} v ,
\end{equation}
with $q_\theta > 0$ additional degree of freedom, the time derivative
in~\eqref{eq:lyapunov1_der} is negative semidefinite; 
using La Salle and Krasowskii principles, asymptotic stability of the
equilibrium point $(y_d,\, \theta_d) = (0,\, 0)$ can therefore be
established, with the \cWalker steadily moving toward the path.

As a consequence, the sign of $\omega$ rules the direction of
switching: a) if $\omega > t_\omega$ then the user has to turn left;
b) if $\omega < -t_\omega$ then the user has to turn right; c) if
$\omega\in [-t_\omega,\, t_\omega]$ then the user has to go straight.
$t_\omega$ is a design threshold used to be traded between the user
comfort and the authority of the control action.

In order to implement the idea of the virtual corridor around the
path and to increase the user comfort, the actuation takes place only when $V_1$
in~\eqref{eq:lyapunov1} is greater than a certain $V_1^{max}$, which
is defined as in~\eqref{eq:lyapunov1} when $y_d = y_h$ is half the
width of the corridor, and $\theta_d = \theta_h$ defines half of the
amplitude of a cone centered on the corridor orientation in which the
\cWalker heading is allowed.  For the haptic, acoustic algorithms the
parameters that define the corridor are the same, that are $y_h =
0.3$~m and $\theta_h = 0.52$~rad.  Similarly, the constants
$q_\theta$, $k_y$ and $k_{\theta}$ are fixed to the same values for
both haptic guidance and acoustic guidance.  However, they are
changing according to the \cWalker actual position: when the position
is outside the corridor, $k_y = 1$ and $k_{\theta} = 0.1$, so that the
controller is more active to steer the vehicle inside the corridor;
when, instead, the \cWalker is within the corridor boundaries, $k_y =
0.1$ and $k_{\theta} = 1$ in order to highly enforce the current
orientation tangent to the path.  The position of the \cWalker inside
the corridor is determined by simply checking if $y_d \leq y_h$.

Finally, to take into account the corridor, the $\omega = \alpha
\omega_d$, where $\alpha$ is a time varying parameter related to the
corridor, i.e., 
\begin{equation}
\label{eq:alpha}
\alpha = min(1, \frac{V_1}{V_1^{max}}) .
\end{equation}
The turning rule related to the sign of $\omega$ is then applied as
previously described.

\subsubsection{Acoustic source computation}
\label{subsubsec:sound}

For the acoustic guidance system, the sound source position has to be
properly identified.  To this end, let us define the circle centered
in the vehicle position $Q$ and having radius $ds$ ($ds = 1.2 m$ in
the experiments). Let us further define $dp$ as the segment joining
the origin of the Frenet-Serret reference frame $F_a$ with the intersection
point $P$ between the circle and the tangent to the circle in the
origin of $F_a$.
If multiple solutions exist, the one being in the forward direction of
the walker is considered.  If only a solution exists, i.e., $y_d =
|ds|$, then $P$ coincides with the origin of $F_a$.  Finally, no
solution exists if $y_d \geq |ds|$, therefore $P$ lies on the segment
that connects $Q$ to $F_a$.

We define $S$ as the point closer to $P$ and lying on the path.  If
the \cWalker is close to a straight component of the path or at a
distance greater than $ds$, $S = P$.  Otherwise, $S$ is computed as
the projection of $P$ on the path.  $S$ is the desired sound source
with respect to a fixed reference frame, which has to be transformed
in the \cWalker reference coordinate systems by
\[
S_{cw} = \begin{bmatrix}
  s_{x_{cw}} \\
  s_{y_{cw}}
\end{bmatrix} =
\begin{bmatrix}
\cos(\theta) & \sin(\theta) \\
-\sin(\theta) & \cos(\theta) 
\end{bmatrix}
(S - Q) .
\]
With the choice just illustrated, if the distance from the path is
greater than $y_d \geq |ds|$, the target is pushed to the planned path
following the shortest possible route.

\subsubsection{Actuation}
\label{subsubsec:actuation}

{\bf Haptic}: The bracelets are actuated according to the direction to
follow.  There are two choices of actuation: the first considers the
value of $\omega$ as discussed above, while the second considers
the value of $s_{y_{cw}}$ as computed in
Sec.~\ref{subsubsec:sound}. In both cases, the sign determines the
direction of turning.

\noindent {\bf Left/Right} Three cones, having the vertices in $Q$,
are defined: $\mathcal{L}$, $\mathcal{R}$ and $\mathcal{S}$.  The
cones divide the semicircle in front of the vehicle and with center in
$Q$ in three equal sectors.  If $S_{cw}\in \mathcal{L}$, the user has
to turn left; if $S_{cw}\in \mathcal{R}$ , the user has to turn right;
if $S_{cw}\in \mathcal{S}$, the user has to go straight.  Positions
behind the user are transformed in turn left or right depending on the
position of $S_{cw}$.  Using this taxonomy and the value of $\alpha$
in~\eqref{eq:alpha}, the slave application determines whether the
sound has to be played or not and from which position.

\noindent {\bf Binaural} The binaural algorithm fully exploits the
reference coordinates $S_{cw}$ using a finer granularity of positions
then the Left/Right acoustic guidance.  The number of cones is now
equal to 7, with three equally spaced cones on the right and on the
left the forward direction of the trolley.  Each cone has a
characteristic angle $\beta_i$, that is the one that equally splits
the cone from the cart perspective.  The described mechanism has the
role of discretising the possible sound directions, since the human
auditory system does not have the sensibility to distinguish a finer
partition.  Again, positions behind the user are treated as in the
Left/Right approach in order to avoid front/back confusion that commonly
affects binaural sound recognition.
As a result of this quantisation, the new
position of the sound source is $S_s$.  By defining with $\theta_i$
the user's head orientation measured with the IMU placed on top of the
headphone, the final sound source $S_p$ is computed as
\[
S_p
=
\begin{bmatrix}
\cos(\theta_i) & \sin(\theta_i) \\
-\sin(\theta_i) & \cos(\theta_i) 
\end{bmatrix}
S_s .
\]

\subsection{Mechanical system: steering}
\label{subsec:mech}

The rationale of the steering wheels controller is in the nature
of the kinematic model.  With respect to the model adopted
in~\eqref{eq:UniModel}, which represents a unicycle-like vehicle model
with differential drive on the back wheels, controlling the \cWalker
using the steering wheels implies a different dynamic for the
orientation rate, that becomes
\begin{equation}
\label{eq:SteeeringAngle}
\dot \theta_d = \frac{\tan(\phi)}{L} v,
\end{equation}
where $\phi$ is the steering angle.  Since the steering angle is
generated through the actuation of the front caster wheels, it is
directly controlled in position by means of the stepper motors.
Moreover, $\phi\in[-\pi,\, \pi)$ and, hence, there is no theoretical
limit on the value of $\dot \theta_d$.  Nonetheless, there exists a
singular point when $v = 0$, which can be ruled out because in such a
case the path following does not have any sense even for the model
in~\eqref{eq:UniModel}.  This condition implies that acting on the
steering wheels do not allow a turn on the spot.

As a consequence, it is possible to select any feasible path following
controller conceived for the unicycle to solve the problem at hand.
The controller adopted is the one proposed by~\cite{SoetantoLP2003},
which is flexible (indeed, there are tuning parameters for the
approaching angle to the path), and can be extended to include dynamic
effects and uncertain parameters.  The adopted controller, which is an
extension of the one presented in~\cite{micaelli1993trajectory}, is
based on the idea of a virtual target travelling on the path. Its
adaptation to our context is discussed below.

Let $V$ be the coordinates of the Virtual vehicle. The objective is to
make the \cWalker perfectly track the Virtual vehicle.  The
position of the walker Q can be expressed in a global frame \textbf{G}
with $^\textbf{G}Q = |Q\ 0|^T = |x\ y\ 0|^T$. Alternatively, the point
can be expressed in the frame \textbf{V}, which coincides with the
point $V$, with $^\textbf{V}Q = |s_v\ y_v\ 0|^T$.

Expressing with $\theta_c$ the orientation in the global frame of the
tangent to the path in $V$, with $s$ the curvilinear abscissa of $V$
along the path, and with $c(s)$ the curvature of the path in that
point, we have $\omega_c = \dot{\theta_c} = c(s)\dot{s}$.  It is now
possible to express the new kinematic equations of the \cWalker w.r.t.
the new frame $V$ starting from~\eqref{eq:UniModel}.  Indeed, starting
from the rotation matrix relating \textbf{G} to \textbf{V}, i.e.,
\[
\begin{aligned}
^\textbf{V}R_{\textbf{G}} =
\begin{bmatrix}
\cos(\theta_c) & \sin(\theta_c) & 0 \\
-\sin(\theta_c) & \cos(\theta_c) & 0 \\
0 & 0 & 1
\end{bmatrix}
\end{aligned} ,
\]
we first derive the velocity of the \cWalker:
\[
^\textbf{G}\dot{Q} = ^\textbf{G}\dot{V} + ^\textbf{G}R_{\textbf{V}}\ ^\textbf{V}\dot{Q}
+ ^\textbf{G}R_{\textbf{V}} (\omega_c \times ^\textbf{V}Q) ,
\]
\[
^\textbf{V}R_{\textbf{G}}\ ^\textbf{G}\dot{Q} = ^\textbf{V}\dot{V} +  ^\textbf{V}\dot{Q}
+ (\omega_c \times ^\textbf{V}Q) ,
\]
\[
^\textbf{V}R_{\textbf{G}}\
\begin{aligned}
\begin{bmatrix}
\dot{x} \\
\dot{y} \\
0
\end{bmatrix}
=
\begin{bmatrix}
\dot{s} \\
0 \\
0
\end{bmatrix}
+
\begin{bmatrix}
\dot{s_v} \\
\dot{y_v} \\
0
\end{bmatrix}
+
\begin{bmatrix}
-c(s)\dot{s}y_v \\
c(s)\dot{s}s_v \\
0
\end{bmatrix}
\end{aligned} ,
\]
and, then, we can express them as
\begin{eqnarray}  
  \begin{bmatrix}
    \dot{s_v} \\
    \dot{y_v} \\
    \dot{\theta_v} 
 \end{bmatrix} = 
  \begin{bmatrix}
   -\dot{s}(1 - c(s)l) + v\cos(\theta_v) \\
   -c(s)\dot{s}s_v + v\sin(\theta_v) \\
   \omega - c(s)\dot{s}
\end{bmatrix} \label{eq:kinematic_virtual}
\end{eqnarray}
where $\theta_v = \theta_d - \theta_c$.  Using Lyapunov techniques, it
is possible to define the following set of control laws
(see~\cite{SoetantoLP2003} for further details)

\[
\dot{s} = v\cos(\theta_v) + K_2 s_v ,
\]
\[
\delta(y_v,v) = -K_{\delta}\tanh(y_vv) ,
\]
\[
\begin{split}
\dot{\theta}_v = &\dot{\delta}(y_v,v) - K_4 K_1 y_v v
\frac{\sin(\theta_v) - \sin(\delta(y_v,v))}{\theta_v - \delta(y_v,v)}
\\&- K_3(\theta_v - \delta(y_v,v)) ,
\end{split}
\]
where $\dot{s}$ represents the progression of the Virtual vehicle on
the path, $\delta(y_v,v)$ is the angle of approach of the vehicle with
respect to the path (that can be tuned as necessary), while the $K_i$
are tuning constants.  With this choice, $\dot{\theta}_v +
c(s)\dot{s}$ is the angular velocity reference $\dot{\theta}_v$ of the
\cWalker, which can be generated by solving with respect to $\phi$ the
equation~\eqref{eq:SteeeringAngle}.

It has to be noted that $\phi$ is the reference of the wheel if the
half-car model is adopted.  In order to transform the reference $\phi$
to a reference for the left and right wheel, the constraint imposed by
the Ackerman geometry are imposed.  Finally, in order to implement the
idea of the corridor previously presented, the actual steering angle
imposed to the wheels considers the reference computed as described in
this section as a reference $\phi_d$, which is used in combination
with the actual orientation $\phi_a$.  More precisely, using the value
of $\alpha$ in~\eqref{eq:alpha}, the commanded orientation of the
steering wheel is given by $\phi = \alpha \phi_d + (1 - \alpha)
\phi_a$.  The value of the threshold in this case is increased to
$\theta_h = 1.62$ rad.

\section{Experimental results}
\label{sec:experiments}

\subsection{Study 1}
%\label{subsec:material}
A formative evaluation was designed to compare and contrast the performance
of the different guidance systems. Since the preliminary state of user research in 
this field~\cite{wilkinson2014demonstrating,wilkinson2014applying}, the main
focus of the evaluation was on system performance, rather 
than on the user experience.
The study had two concurrent objectives: to  develop a controlled experimental
methodology to support system comparisons and to provide practical information
to re-design.
In this way, future development will provide a tested methodology to preserve
elderly participants from stress and fatigue.
In line with an ethical  application of the inclusive design
process~\cite{keates2003countering}, at this early stage of the methodological
verification process of an evaluation protocol, we involved a sample of
young participants.

\subsubsection{Participants}
Thirteen participants (6 females, mean age 30 years old, ranging from 26 to 39)
took part in the evaluation. They were all students or employees of the
University of Trento and gave informed consent prior to inclusion in the study.

\subsubsection{Design}
The study applied a within-subjects design with Guidance~(4) and Path~(3) as 
experimental factors. All participants used the four guidance systems
(acoustic, haptic, mechanical, binaural) in three different paths.
The order of the system conditions was counterbalanced across participants.

\subsubsection{Apparatus}
The experimental apparatus used in the experiment is a prototype of
the \cWalker shown in Figure~\ref{fig:cWalker}.  An exaustive
description of the device and of its different functionalities can be
found in~\cite{pal15}.  A distinctive mark of the \cWalker is its
modularity: the modules implementing the different functionalities can
be easily plugged on or off based on the specific
requirement of the application.  The specific configuration adopted in
this paper consisted of: 1. a {\tt Localisation} module, 2. a short
term {\tt Planner}, 3. a {\tt Path Follower}.

The {\tt Localisation} system of the \cWalker utilises a combination of
different techniques. A relative localisation system based on the
fusion of encoders on the wheels and of a multi-axial gyroscope,
operates in connection with different absolute positioning systems to
keep in check the error accumulated along the path.  The experiments
reported here were organised as multiple repetitions of relatively short
trajectories. We believe that the adoption of this paradigm produces
results comparable to ``fewer'' repetitions of longer trajectories, in
a more controllable and repeatable way.  This simplifies the
localisation problem. Indeed, the mere use of relative localisation provides
acceptable accuracy with an accumulated error below $5cm$, when the
system operates for a small time (e.g., smaller than
$50$m)~\cite{NazemzadehFMP14}. Therefore the activation of absolute positioning
systems which would entail some instrumentation in the environment (e.g.,
by deploying RFID tags in known positions) was not needed.

The short term {\tt Planner} in the \cWalker is reactive: it collects real--time
information in the environment and uses it to plan safe
courses that avoid collisions with other people or dangerous areas~\cite{ColomboFLPS13}. In this context, we could disable this
feature since the experiments took place in free space, without any
dynamic obstacles along the way. The planner was configured to
generate three different virtual paths (60 centimetres wide and 10
meters long): straight (I), C shaped (C) and S shaped (S). The
width of the virtual corridor was above 30 centimetres to the left and to the
right of centre of the \cWalker. The C path was a quarter of
the circumference of a circle with a radius of 6.37 meters. The S path
comprised three arches of a circumference with a radius of 4.78
meters. The first and the third arches were $\sfrac{1}{12}$, while the
one in the centre was $\sfrac{1}{6}$ of the whole circumference. The
second arch was bent in the opposite direction compared to the other
two. In total there were 6 path variations, two symmetric paths for
each shape.

Finally, the {\tt Path Follower} component implements the guidance
algorithms described in Section~\ref{sec:systems}. The concrete
implementation was adapted to the different guidance algorithms.  For
mechanical guidance, the component decides a direction for the wheel
that is transmitted to a {\tt Wheel Position Controller}. This
component also receives real--time information on the current
orientation of the wheel and decides the actuation to set the
direction to the desired position.  For the haptic and the acoustic
(and binaural) guidance, the {\tt Path Follower} implements the
algorithms discussed in Section~\ref{sec:haptic_sound} and transmits
its input either to the {\tt Haptic Slave} or to the {\tt Audio
  Slave}, as detailed in Section~\ref{subsec:bracelets} and
Section~\ref{subsec:audio_interface} respectively.

\subsubsection{Procedure}
%\label{subsec:procedure}

The evaluation was run in a large empty room of the University building by
two experimenters: a psychologist who interacted with the participants and
a computer scientist who controlled the equipment. At the beginning of the
study, participants were provided with the instructions in relation to each
guidance system. It was explained that they had to follow the instruction of
the \cWalker: while they were on the correct trajectory there would be no
system intervention. Otherwise, each system would have acted in different ways.
The mechanical system would have turned the front wheels modifying its
direction onto the right path. In this case, participants could not force the
walker and might only follow the suggested trajectory. At the end of the
mechanical correction, the participants were given back the control of the
walker. For the haptic/acoustic guidance, a vibration/sound (either on the
left or right arm/ear) would have indicated the side of the correction
necessary to regain the path. It was stressed that under these conditions there
was no information indicating the turn intensity. Finally, the binaural
guidance would have provided a sound indicating the direction and (the amount
of the correction).

Participants were told to be careful in following the instructions to avoid
bouncing from one side to the other of the virtual corridor. It was also
suggested that whenever they felt like zigzagging, the actual trajectory
might be likely in the middle.

Before each trial, the appropriate device was put on the participant (i.e.,
headphones or haptic bracelets). Only in the case of the binaural system,
participants were given a brief training to make them experience the spatial
information of the sounds. The starting position of each trial varied among the
four corners of a rectangular virtual area (about 12 x 4 meters). The \cWalker
was positioned by the experimenter with a variable orientation
according to the shape of the path to be followed. Specifically, at the
beginning of each I trial, the walker was turned 10 degrees either to the left
or to the right of the expected trajectory. At the beginning of each C and S
trials, the walker was located in the right direction to be followed.
Participant started walking after a signal of the experimenter and
repeated 10 randomised paths for each guidance system.

At the end of each system evaluation, participants were invited to
answer 4 questions, addressing ease of use, self-confidence in route keeping,
acceptability of the interface in public spaces and an overall evaluation
on a 10 points scale (10=positive). Moreover, participants were invited to
provide comments or suggestions. The evaluation lasted around 90 minutes,
at the end participants were thanked and paid 10 Euros.

\subsubsection{Data analysis}
%\label{subsec:analysis}

Performance was	analysed considering four dependent variables. A measure of
error was operationalized as deviation from the optimal trajectory and
calculated using the distance of the orthogonal projection between the actual
and the optimal trajectory. We collected a sample of 100 measurement
(about one value every 10 centimetres along the curvilinear abscissa of the
path) that were then averaged. \textit{Time} was measured between the start of
participant's movement and the moment the participant reached the intended
end of the path. \textit{Length} measured the distance walked by the
participant, whereas \textit{speed} corresponded to the ratio between the
length and the time. 

For each participant and guidance system, we averaged an index scores
for the four S, the four C and the two I paths. Data analysis was
performed employing the analysis of variance (ANOVA) with repeated measures
on the factors `Guidance' and `Path'. Post-hoc pairwise comparisons corrected
with Bonferroni  for multiple comparisons (two tails) were also computed.

\subsubsection{Results}

\begin{figure*}[t]
\centering
\begin{tabular}{cc}
  \includegraphics[width=0.45\textwidth]{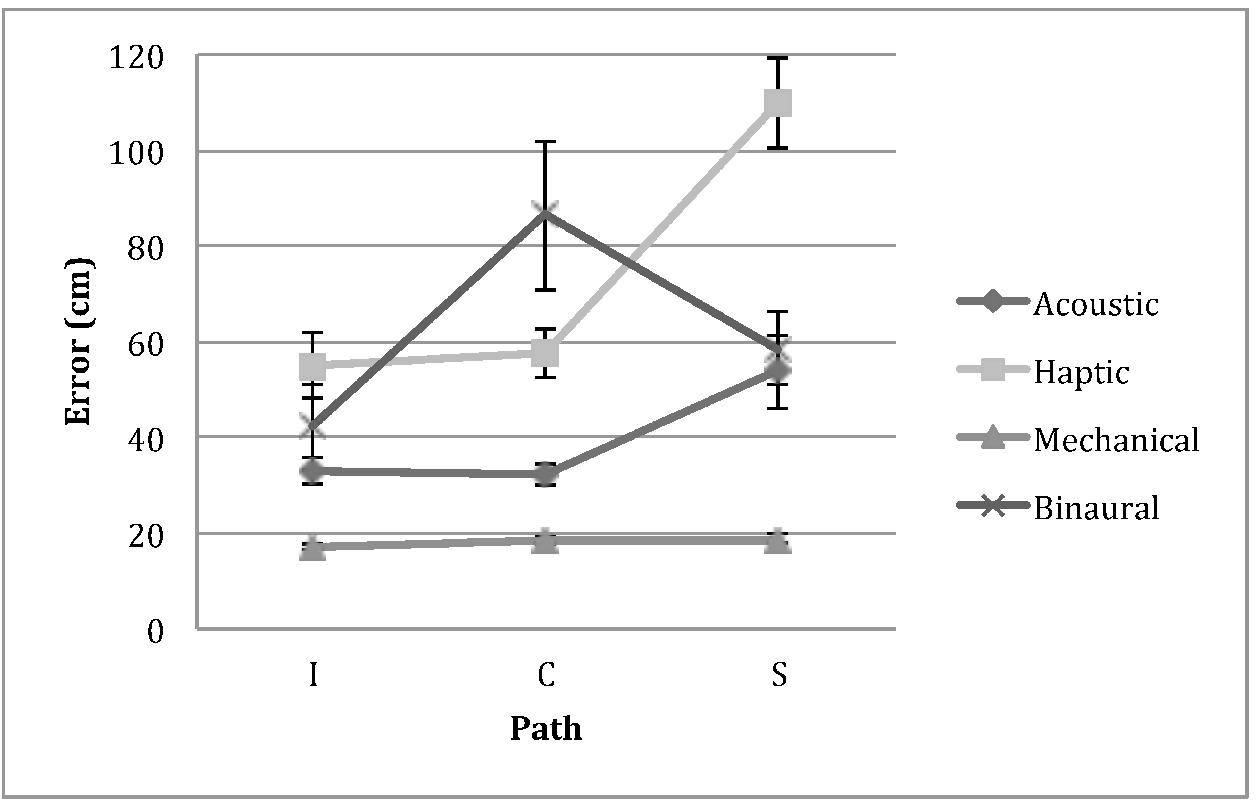} &
  \includegraphics[width=0.45\textwidth]{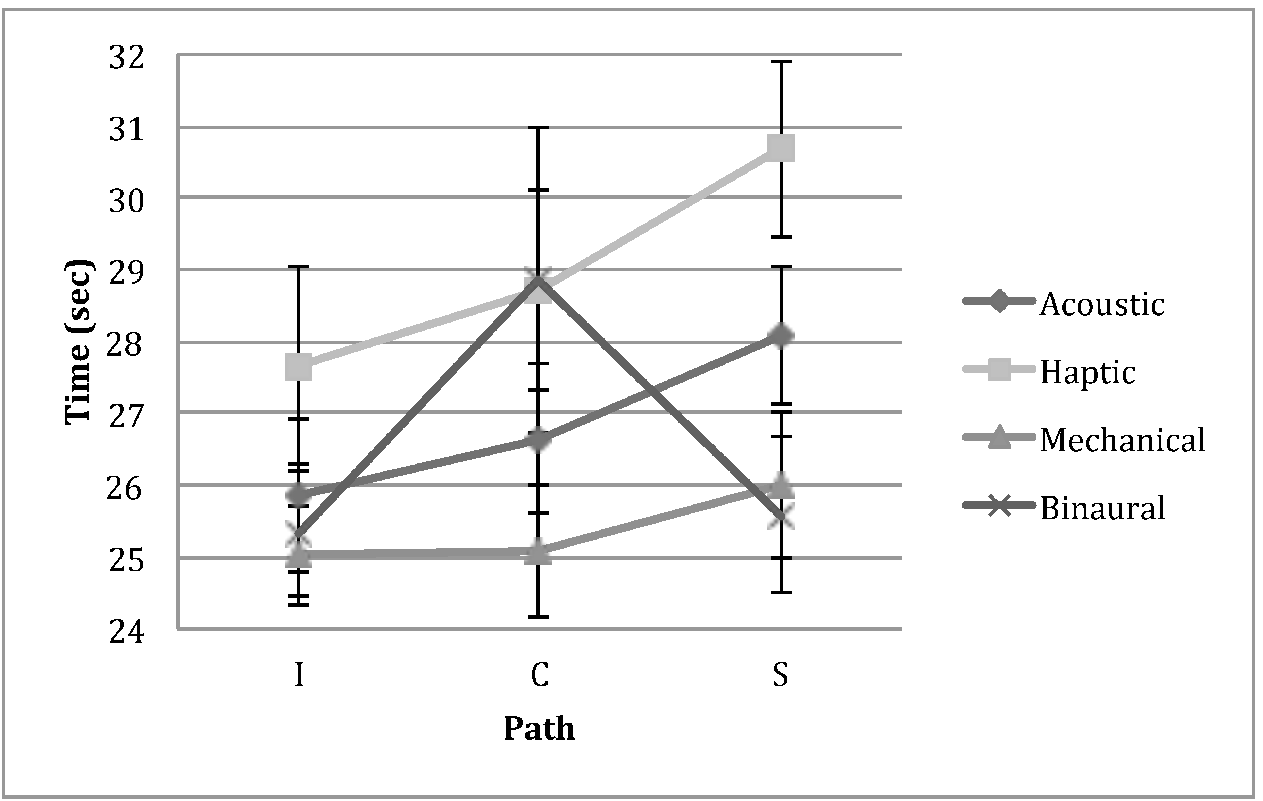} \\
  (a) &
  (b) \\
  \includegraphics[width=0.45\textwidth]{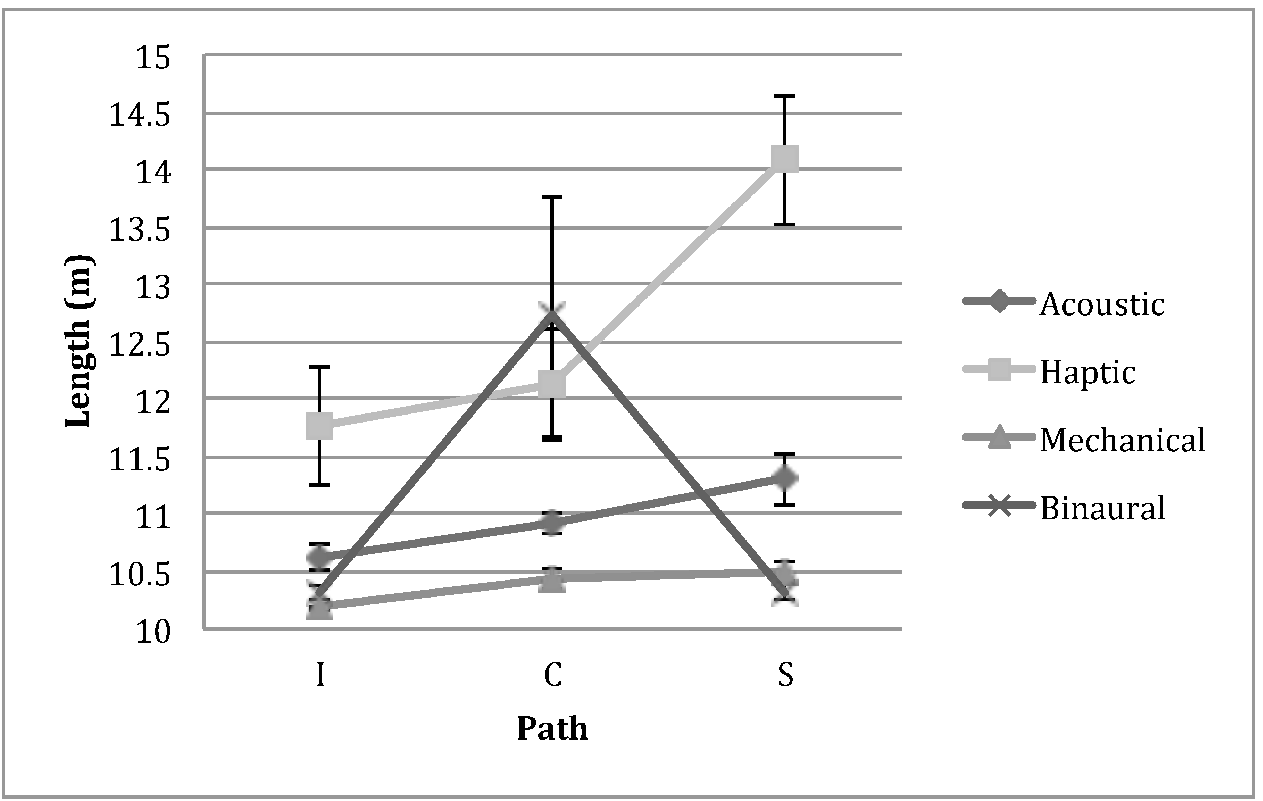} &
  \includegraphics[width=0.45\textwidth]{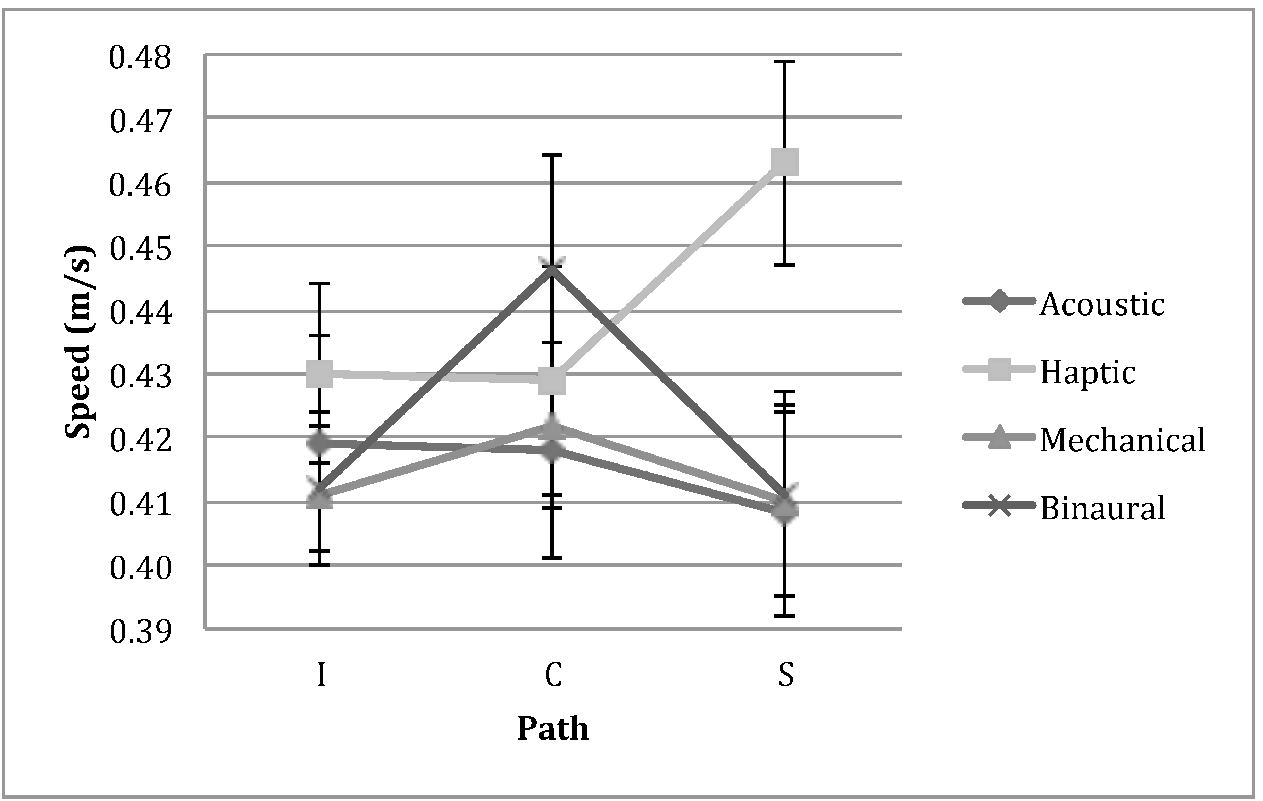} \\
  (c) &
  (d) \\
\end{tabular}
\caption{Different metrics as function of Guidance and Path:
(a) average error (cm),
(b) average time (s),
(c) average length (m),
(d) average speed (m/s).}
\label{fig:exp_study1}
\end{figure*}

\noindent {\bf Error}
Descriptive statistics of error are reported in Fig.~\ref{fig:exp_study1}~(a) as a
function of Guidance and Path. The ANOVA highlighted a significant effect for
Guidance $F (3,36) = 27.4$, $p < .01$, Path $F (2,24) = 17.3$, $p < .01$ and
for the interaction $F (6,72) = 10.3$, $p < .01$. Post-hoc pairwise comparison
(Tab.~\ref{tab:exp_tab1}) indicated that the mechanical guidance differed
significantly from all the others ($p < .01$) being the most precise. Moreover,
the acoustic guidance was significantly different from the haptic ($p < .01$).
Post-hoc comparisons indicated that the I path was significantly easier from
the other two ($p < .01$).

\begin{figure*}[t]
\centering
\begin{tabular}{cc}
  \includegraphics[width=0.45\textwidth]{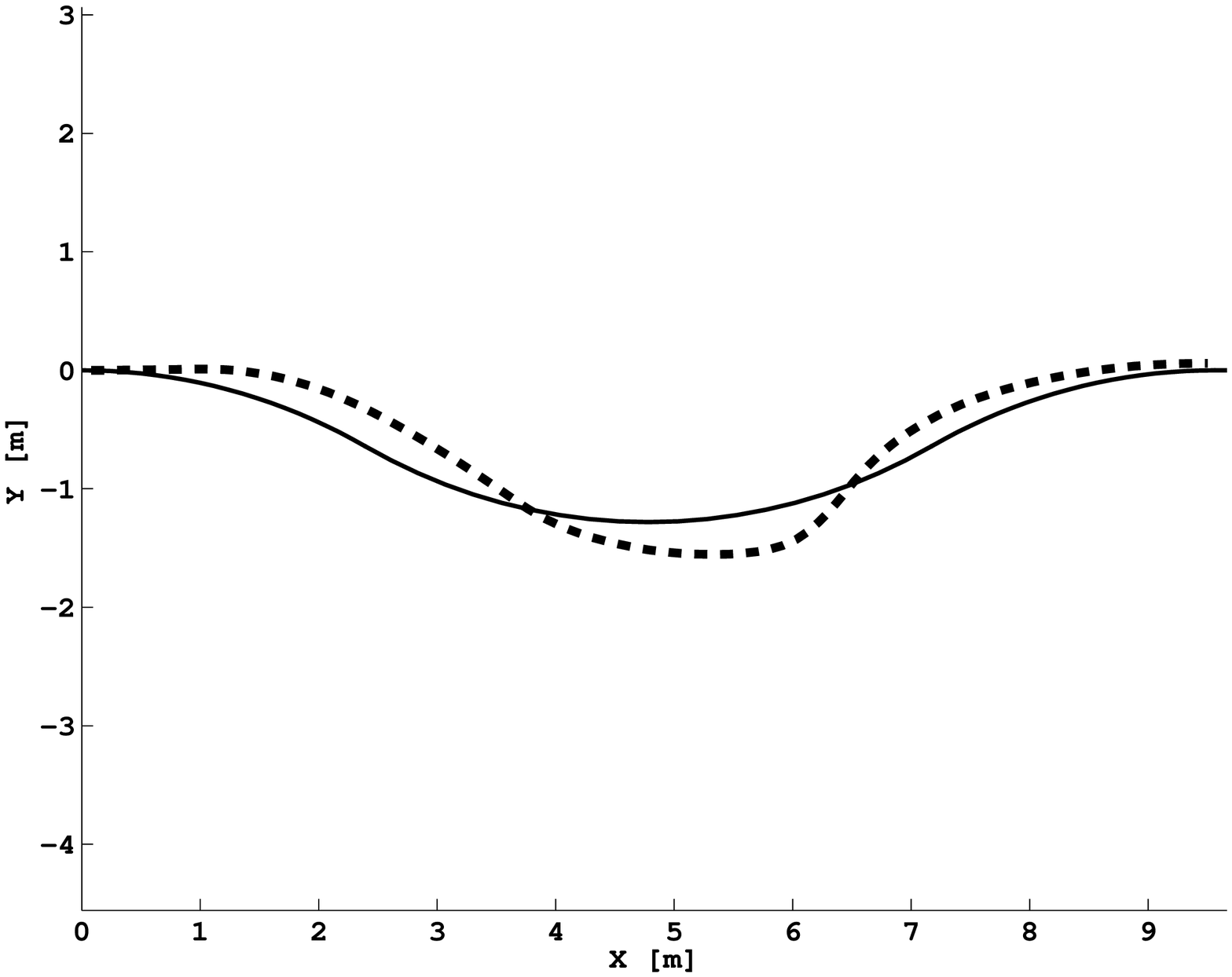} &
  \includegraphics[width=0.45\textwidth]{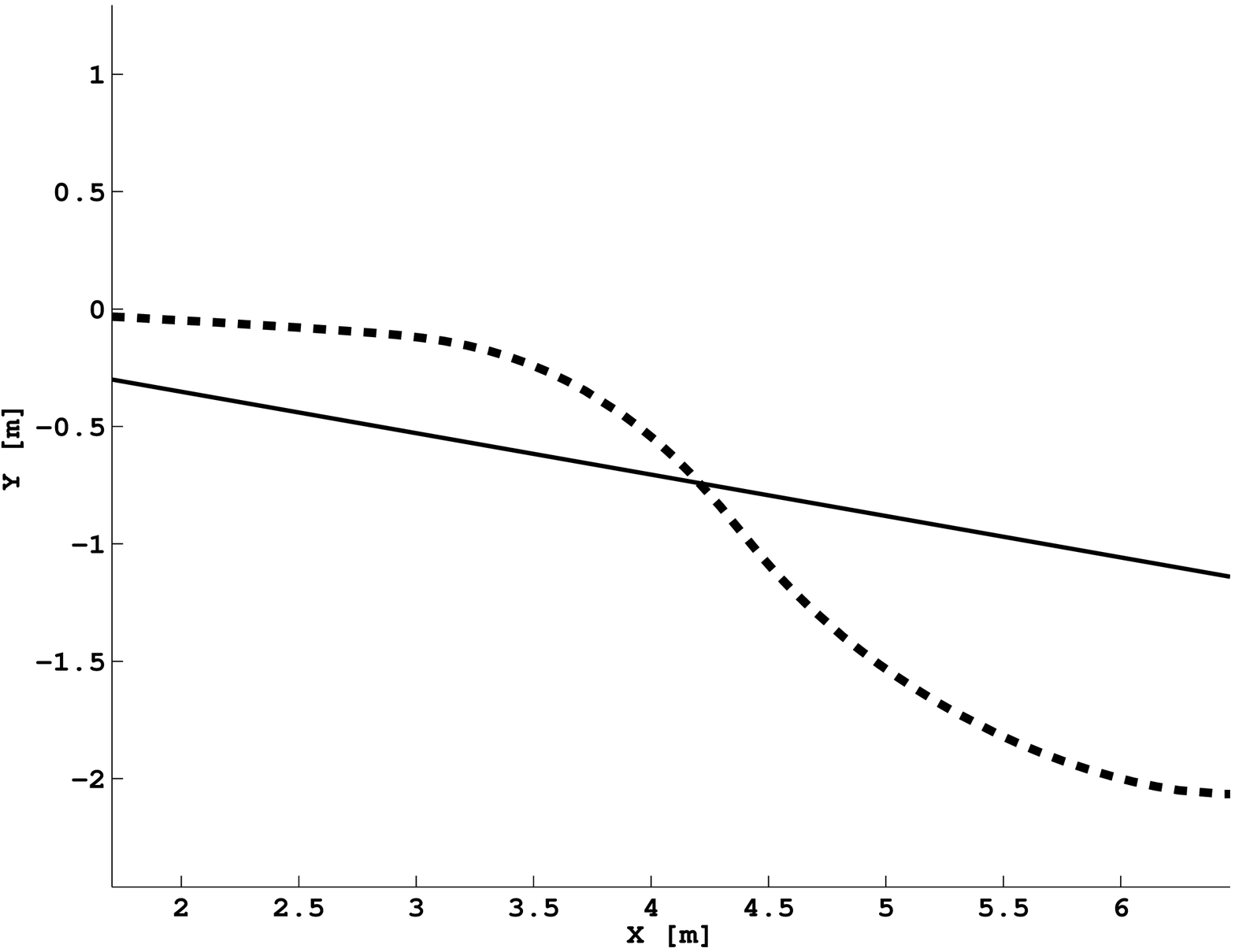} \\
  (a) &
  (b) \\
  \includegraphics[width=0.45\textwidth]{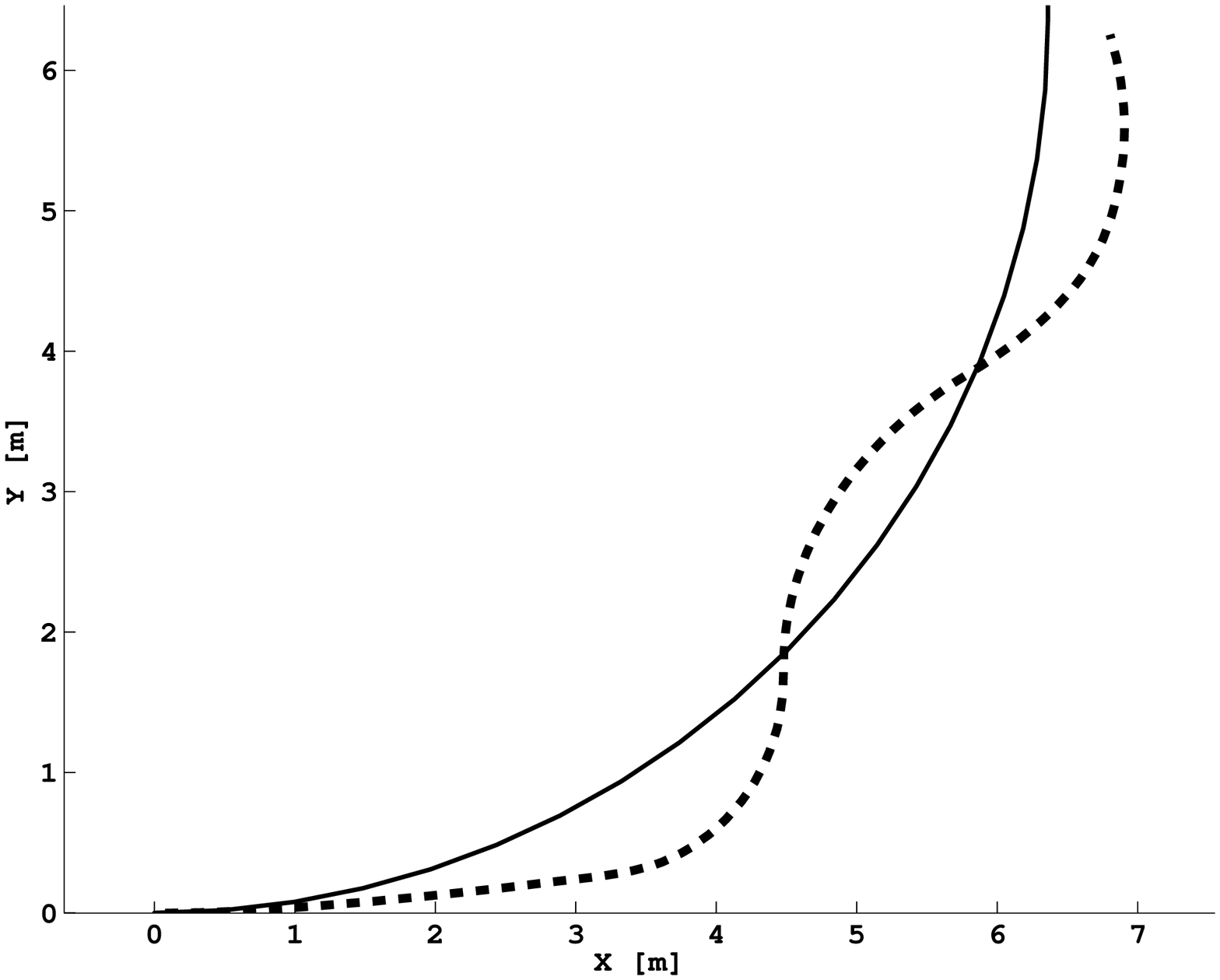} &
  \includegraphics[width=0.45\textwidth]{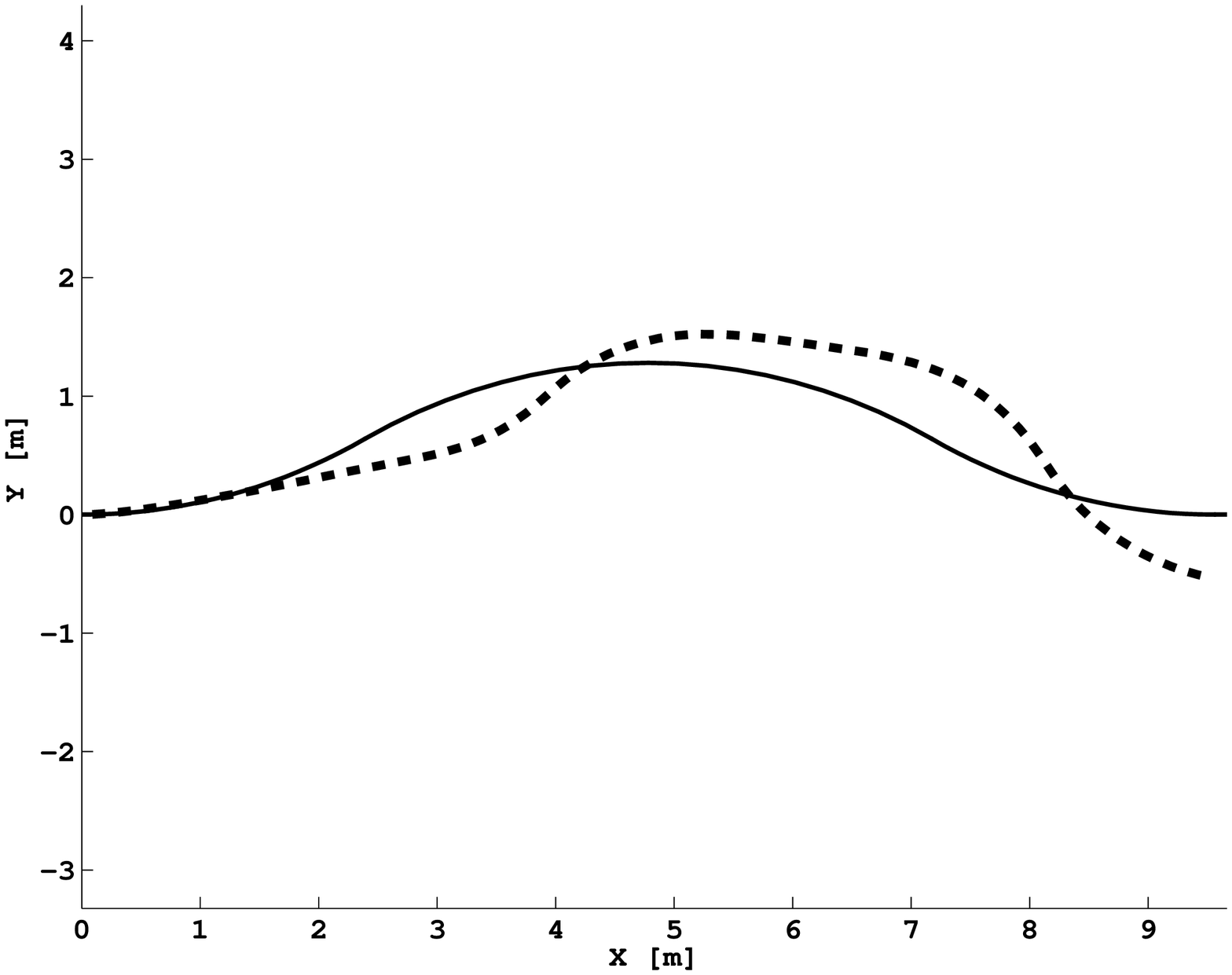} \\
  (c) &
  (d) \\
\end{tabular}
\caption{Some examples of qualitative results for different guidance algorithms:
(a) mechanical,
(b) haptic,
(c) audio,
(d) binaural.
The trajectory of the user is the dash line.}
\label{fig:exp_qual}
\end{figure*}

In the mechanical guidance condition, the error was not affected by the path
and showed very low variability among participants. On the contrary, for all
other conditions there was an effect of Path on the magnitude of the error.
Mostly for the haptic, but also for the acoustic guidance, the S path had the
highest error. Interestingly, for the binaural guidance, the highest error
emerged with the C path.
Fig.~\ref{fig:exp_qual} shows some qualitative results of the experiments.

\begin{table*}
\begin{tabularx}{\textwidth}{|C|C|C|C|C|C|}
\hline
\textbf{Guidance} & \textbf{I} & \textbf{C} & \textbf{S} & \textbf{Average}
& \pbox{5cm}{\textbf{Pairwise}\\\textbf{comparison}} \tabularnewline
\hline
\textbf{Acoustic} & 33.0 & 32.2 & 53.7 & \textbf{39.6}
& I vs. S, $p = .053$ C vs. S, $p < .05$ \tabularnewline
\hline
\textbf{Haptic} & 55.0 & 57.4 & 109.7 & \textbf{74.0}
& I vs. C and S, $p < .01$ \tabularnewline
\hline
\textbf{Mechanical} & 17.1 & 18.3 & 18.8 & \textbf{18.1} & n.s. \tabularnewline
\hline
\textbf{Binaural} & 42.1 & 86.3 & 58.7 & \textbf{62.4}
& I vs. C, $p < .05$ \tabularnewline
\hline
\textbf{Average} & \textbf{36.8} & \textbf{48.6} & \textbf{60.2} & & \tabularnewline
\hline
\pbox{5cm}{\textbf{Pairwise}\\\textbf{comparison}}
& M vs. A and H, $p < .01$ A vs. H, $p = .07$
& A vs. H and M, $p < .01$ A vs. B, $p < .05$ M vs. H and B, $p < .01$
& M vs. all, $p < .01$, H vs. A and B, $p < .01$ & & \tabularnewline
\hline
\end{tabularx}
\caption{Average error (cm) for each experimental condition and significant
post-hoc pairwise comparisons.}
\label{tab:exp_tab1}
\end{table*}

\noindent {\bf Time}
The ANOVA on time showed a significant effect for Guidance $F (3,36) = 3.98$,
$p < .05$, Path $F (2,24) = 7.54$, $p < .01$ and for the interaction
$F (6,72) = 2.89$, $p < .05$. Fig.~\ref{fig:exp_study1}~(b) shows the average time
in relation to both Guidance and Path. Post-hoc pairwise comparison showed that
the mechanical guidance system was significantly faster than the haptic
($p < .05$), and that the I path differed significantly from the S path
($p < .05$). Walking time was independent of Path for the mechanical and the
binaural guidance. Conversely, the S path was performed significantly slower
than the I path for both the acoustic and the haptic guidance. The average time
and the results of the post-hoc pairwise comparisons are summarized in
Tab.~\ref{tab:exp_tab2}. 

\begin{table*}
\begin{tabularx}{\textwidth}{|C|C|C|C|C|C|}
\hline
\textbf{Guidance} & \textbf{I} & \textbf{C} & \textbf{S} & \textbf{Average}
& \pbox{5cm}{\textbf{Pairwise}\\\textbf{comparison}} \tabularnewline
\hline
\textbf{Acoustic} & 25.8 & 26.7 & 28.1 & \textbf{26.9}
& I vs. S, $p < .01$ \tabularnewline
\hline
\textbf{Haptic} & 27.7 & 28.7 & 30.7 & \textbf{29.0}
& I vs. S, $p < .05$ \tabularnewline
\hline
\textbf{Mechanical} & 25.0 & 25.1 & 26.0 & \textbf{25.4} & n.s. \tabularnewline
\hline
\textbf{Binaural} & 25.3 & 28.9 & 25.6 & \textbf{26.6} & n.s. \tabularnewline
\hline
\textbf{Average} & \textbf{26.0} & \textbf{27.3} & \textbf{27.6} & & \tabularnewline
\hline
\pbox{5cm}{\textbf{Pairwise}\\\textbf{comparison}}
& n.s.
& n.s.
& H vs. M and B, $p < .05$ & & \tabularnewline
\hline
\end{tabularx}
\caption{Average time (sec) for each experimental condition and significant
post-hoc pairwise comparisons.}
\label{tab:exp_tab2}
\end{table*}

\noindent {\bf Length}
Only two participants (both in the S path and the binaural guidance condition)
walked less than the optimal path length. The ANOVA showed a significant
effect for Guidance $F (3,36) = 15.1$, $p < .01$, Path $F (2,24) = 9.1$,
$p < .01$ and for the interaction $F (6,72) = 6.1$, $p < .01$. Post-hoc
comparisons indicated that the haptic guidance differed significantly from all
the others ($p < .01$ mechanical and acoustic and $p < .05$ binaural).
Moreover, the mechanical guidance differed significantly from the acoustic
($p < .01$). The I path differed significantly from the C ($p < .01$) and S
($p < .05$) paths (Tab.~\ref{tab:exp_tab3}). The haptic guidance showed the
worst result in the S path. For the mechanical condition, the performance was
different between the I and S path. For the binaural condition there was no
effect of Path.
Fig.~\ref{fig:exp_study1}~(c) shows the average length in relation to both
Guidance and Path.

\begin{table*}
\begin{tabularx}{\textwidth}{|C|C|C|C|C|C|}
\hline
\textbf{Guidance} & \textbf{I} & \textbf{C} & \textbf{S} & \textbf{Average}
& \pbox{5cm}{\textbf{Pairwise}\\\textbf{comparison}} \tabularnewline
\hline
\textbf{Acoustic} & 10.62 & 10.92 & 11.3 & \textbf{10.9}
& \pbox{5cm}{I vs. C, $p < .05$\\I vs. S, $p < .01$} \tabularnewline
\hline
\textbf{Haptic} & 11.76 & 12.12 & 14.08 & \textbf{12.7}
& \pbox{5cm}{S vs. I, $p < .01$\\S vs. C, $p < .05$} \tabularnewline
\hline
\textbf{Mechanical} & 10.18 & 10.44 & 10.5 & \textbf{10.4}
& I vs. S, $p < .01$ \tabularnewline
\hline
\textbf{Binaural} & 10.31 & 12.72 & 10.32 & \textbf{11.1} & n.s. \tabularnewline
\hline
\textbf{Average} & \textbf{10.7} & \textbf{11.6} & \textbf{11.6} & & \tabularnewline
\hline
\pbox{5cm}{\textbf{Pairwise}\\\textbf{comparison}}
& M vs. A, $p < .05$
& M vs. H, $p < .05$ M vs. A $p = .07$
& \pbox{5cm}{all the\\comparisons\\$p < .01$ except\\M vs. A, 
$p < .05$\\and M vs. B\\not significant}
& & \tabularnewline
\hline
\end{tabularx}
\caption{Table 3 Average travelled length (m) for each experimental condition
and significant post-hoc pairwise comparisons.}
\label{tab:exp_tab3}
\end{table*}

\noindent {\bf Speed}
The analysis of variance on speed reported only a significant interaction
between Guidance and Path $F (6,72) = 3.05$, $p < .01$.
Fig.~\ref{fig:exp_study1}~(d) reports
the average time as a function of experimental conditions. The average speed
and the results of the post-hoc pairwise comparisons are summarized in
Tab.~\ref{tab:exp_tab4}. Participants were particularly fast walking the S path. 

\begin{table*}
\begin{tabularx}{\textwidth}{|C|C|C|C|C|C|}
\hline
\textbf{Guidance} & \textbf{I} & \textbf{C} & \textbf{S} & \textbf{Average}
& \pbox{5cm}{\textbf{Pairwise}\\\textbf{comparison}} \tabularnewline
\hline
\textbf{Acoustic} & 0.42 & 0.42 & 0.41 & \textbf{0.42} & n.s. \tabularnewline
\hline
\textbf{Haptic} & 0.43 & 0.43 & 0.46 & \textbf{0.44}
& I vs. S, $p < .01$ \tabularnewline
\hline
\textbf{Mechanical} & 0.41 & 0.42 & 0.41 & \textbf{0.41} & n.s. \tabularnewline
\hline
\textbf{Binaural} & 0.41 & 0.45 & 0.41 & \textbf{0.42} & n.s. \tabularnewline
\hline
\textbf{Average} & \textbf{0.42} & \textbf{0.43} & \textbf{0.42} & & \tabularnewline
\hline
\pbox{5cm}{\textbf{Pairwise}\\\textbf{comparison}}
& n.s.
& n.s.
& A vs. H, $p < .05$ M vs. H, $p = .07$ & & \tabularnewline
\hline
\end{tabularx}
\caption{Average speed (m/sec) for each experimental condition and
significant post-hoc pairwise comparisons.}
\label{tab:exp_tab4}
\end{table*}

\subsubsection{Questionnaire}
Participants scores to the the four questionnaire items were normalized for
each participant in relation to the highest score provided among all the
answers.
The ANOVA indicates that the mechanical guidance is perceived as easier to use
with no other significant differences among the other systems. The same
results have emerged in relation to the confidence to maintain the correct
trajectory. Concerning the acceptability to use the guidance systems in
public spaces, the mechanical guidance was again the preferred one in
relation to both the acoustic and binaural while no difference has emerged in
relation to the haptic. Finally, participants liked the
mechanical guidance the most in relation to both haptic and acoustic systems,
while no difference has emerged in relation to the binaural one.

Participants spontaneously commented that the mechanical system was easy to
follow and required little attention. However, some of them complained that
it might be perceived as coercive and risky due to possible errors in route
planning. Other people worried about the dangerous effect of a quick turn of
the wheels mostly for older users. Participants reported a general dislike
about wearing headphones mostly because they might miss important environmental
sounds and because of the look. Most of the participants agreed that the
binaural condition required more attention than all the other systems.
Participants however appreciated that it was something new, interesting and
provided a constant feedback on the position. Most of them preferred the
binaural system to the acoustic one because it provided more information, yet
some reported a difficulty in discriminating the direction the sound was coming. 

Most of the participants reported to prefer the haptic guidance system to the
acoustic, as easier and less intrusive. In relation to the both guidance
condition, participants complained about the poverty of the left and right
instructions and the lack of a modulation. Some participants suggested possible
ways to increase communication richness, such as, for the acoustic system,
different volume indicating the magnitude; verbal feedback; different tones in
relation to the angle. For the haptic system comments included modulating the
frequency of the vibration in relation to the magnitude of the correction.
Some participants reported a kind of annoyance for the haptic stimulation but
only for the first minutes of use.

\subsubsection{Discussion}
The aim of this study was to gather quantitative and qualitative information in
relation to the evaluation of four different guidance systems. To this aim 
participants had the opportunity to navigate non-visible paths (i.e., virtual 
corridors) using four the different guidance systems. To maintain the correct 
trajectory participants could only rely on the instructions provided by the
\cWalker and, after using each system, they were asked to provide feedback.

As expected, in terms of performance, the mechanical guidance was the most 
precise. Although an error emerged because of the freedom left to participants, 
the results show the consistency of the deviation along the different paths, a low 
variability among the participants and a slight difference in relation to the shape 
of the paths. The results of the questionnaire further support quantitative data 
showing that, on average, participants liked the mechanical guidance the most in 
relation to easiness, confidence in maintaining the trajectory, acceptability and 
overall judgment. The only concern for some users was that it might be 
perceived as coercive and risky due to possible errors in route planning. In fact, 
the mechanical guidance was active in the sense that participants had to 
passively follow the trajectory imposed by the walker. Differently, in the other 
three guidance systems, the participants were actively driving based on the 
interpretation of the provided instructions. In the acoustic guidance, there were 
only left and right sounds while in the binaural guidance, the sound was 
modulated by modifying the binaural difference between the two ears. Although 
more informative, in terms of quantifying the angle of the suggested trajectory, 
the binaural guidance system emerged to be worse than the acoustic system in 
the C path. However, it is likely that with adequate training the performance 
with the binaural system could improve a lot. The results of the questionnaire 
suggest that both the systems using headphones were not very acceptable 
because of the possibility to miss environmental sounds and because of the look. 
Moreover, the binaural system was reported to require more attention than the 
acoustic one, although no difference emerged in terms of confidence in 
maintaining the correct trajectory. Overall, the binaural guidance was 
appreciated because it was something new and provided detailed information. 
Indeed, most of the participants' suggestions related to the acoustic and haptic 
guidance systems were addressed at codifying the instructions in terms of the 
angle of the correction. 

Significant performance differences emerged 
between the haptic and the acoustic guidance, which could in part be explained 
by the natural tendency to respond faster to auditory stimuli rather than to 
tactile stimuli, and in part by the different algorithm employed in the evaluation. 
This issue is addressed in the second study presented in this paper. Looking at 
participants performance however it is evident that, independent of the 
communication channel, the dichotomous nature of the stimulation (left-right) 
tended to stimulate long left and right corrections leading to zigzagging. One 
participant explicitly mentioned this feeling while commenting on the haptic 
guidance. In terms of user experience, the haptic guidance was perceived as 
more acceptable than the acoustic and the binaural systems, and no different 
from the mechanical one. Indeed, most of the participants commented that the 
haptic bracelets could be hidden and did not interfere with the environmental 
acoustic information.

\subsection{Study 2}

This evaluation study was designed to clarify the differences emerged in study 1
between the haptic and the acoustic guidance. To this aim both input devices 
(bracelets and headphones) were interfaced to the same guidance algorithm and 
tested following the same experimental protocol as study 1, except that 
participants were required to test only the acoustic and the haptic guidance 
systems. Moreover, the haptic guidance system was modified using the acoustic 
guidance algorithm. In this way, we could test directly the effect of the interface.  
Ten participants (2 females, mean age 30 years old range 24-35) took part in the 
study.

\subsubsection{Results}
\begin{figure}
\includegraphics[width=\columnwidth]{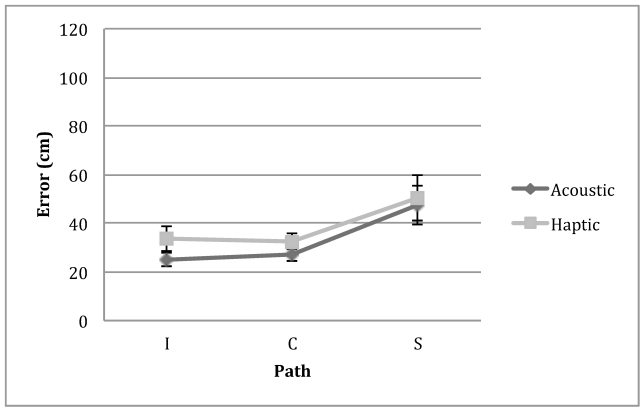}
\caption{Average error (cm) as a function of Guidance and Path.}
\label{fig:exp_fig6}
\end{figure}

Descriptive statistics of error are reported in Fig.~\ref{fig:exp_fig6} as a
function of Guidance and Path. The ANOVA showed a significant effect for the
factor Path $F (2,18) =  11.0$, $p < .01$ but not Guidance. The post-hoc
pairwise comparison showed that the S path differed significantly from the
other two ($p < .05$) confirming its higher complexity.

The ANOVA on time, length, and speed returned the same trend of results: 
a main effect only for Path. The analysis of the questionnaire confirmed a 
preference for the acceptability of the haptic guidance in public spaces. Finally, a 
between-study analysis of variance comparing the performance of participants 
using the sound system in study 1 and study 2 returned no significant 
differences due to study, path, or their interaction.

\subsubsection{Discussion}
The study indicates that the haptic and acoustic interfaces do not differ in
terms of performance, and that the results of study 1 may be attributed
entirely to the different algorithms tested. Furthermore, they confirm a
preference for the haptic guidance but only regarding its social acceptability
in public spaces. 
Furthermore, the similarities in both performance and user-experience of the 
acoustic guidance in the two studies is an indicator of the strong reliability
and external validity of the evaluation protocol.

Tab.~\ref{tab:exp_tab5} propose a ranking of the 4 guidance systems,
combining empirical observations, measurements and participants' comments in
both studies. The best guidance was no doubt the mechanical one, followed
by the haptic, acoustic and binaural systems. The evaluations highlighted
new challenges for the sociotechnical design of future guidance system.
In particular a major issue emerged with regards to the acceptability of the
practical requirement of wearing headphones.
The binaural system was perceived as a promising solution which
captured the user attention. 
However, more work is needed in order to improve the communication of the 
directional information.

\begin{table*}
\begin{tabularx}{\textwidth}{|C|C|C|C|C|C|}
\hline
 & {\bf Performance} & {\bf Easiness} & {\bf Confidence}
& {\bf Acceptability} & {\bf Total} \tabularnewline
\hline
{\bf Acoustic} & ++ & ++ & ++ & + & 7 \tabularnewline
\hline
{\bf Haptic} & ++ & ++ & ++ & +++ & 9 \tabularnewline
\hline
{\bf Mechanical} & ++++ & +++ & +++ & +++ & 13 \tabularnewline
\hline
{\bf Binaural} & + & + & ++ & + & 5 \tabularnewline
\hline
\end{tabularx}
\caption{Summary of the results.}
\label{tab:exp_tab5}
\end{table*}

\section{Conclusions}
\label{sec:conclusions}

In this paper we have presented four different solutions for guiding a user
along a safe path using a robotic walking assistant. One of them is
``active'' meaning that the system is allowed to ``force a turn'' in a
specified direction. The other ones are ``passive'' meaning that they
merely produce directions that the user is supposed to follow on her
own will.  We have described the technological and scientific
foundations for the four different guidance systems, and their implementation
in a device called \cWalker.

The systems has been thoroughly evaluated with a group of young volunteers,
allowing us to test the methodology, and providing a baseline for future
tests with potential elderly users.
This paper contributed a novel evaluation protocol for comparing the different
guidance systems, and opens new challenges for interaction designers.
The use of virtual corridors allowed us to test the precision of the guidance
systems to maintain the correct trajectory in the absence of any visual
indications of the route.
However, in a real-life scenario, users would most likely walk along a wide
corridor with walls on the left and right that might help maintaining a
straight path in a particular part of the corridor (i.e., in the centre or
towards the left/right).
Moreover, corridors' crossings are often orthogonal. In such scenarios, left
and right instructions might be enough to allow the user to reach their goal
and the haptic solution could be the best trade off among precision, freedom
and cognitive workload, leaving vision and audition free to perceive
environmental stimuli.
Future research, will repeat this study in more ecological contexts.

From the technical point of view, an interesting future direction is the
implementation of guidance solutions based on the use of electromechanical
brakes, along the lines suggested in our previous work~\cite{FontanelliGPP13}.

\bibliographystyle{abbrv}
\bibliography{article}  

\begin{thebibliography}{10}

\bibitem{Arias2010}
A.~Arias and U.~Hanebeck.
\newblock Wide-area haptic guidance: Taking the user by the hand.
\newblock In {\em Proc. IEEE/RSJ Int. Conf. Intel. Robots Syst.}, pages
  5824--5829, 2010.

\bibitem{blauert}
J.~Blauert.
\newblock {\em Spatial Hearing-Revised Edition: The Psychophysics of Human
  Sound Localization}.
\newblock MIT press, 1996.

\bibitem{Cacioppo2009}
J.~T. Cacioppo and L.~C. Hawkley.
\newblock Perceived social isolation and cognition.
\newblock {\em Trends in Cognitive Sciences}, 13(10):447 -- 454, 2009.

\bibitem{chuy2006new}
O.~Chuy, Y.~Hirata, and K.~Kosuge.
\newblock A new control approach for a robotic walking support system in
  adapting user characteristics.
\newblock {\em Systems, Man, and Cybernetics, Part C: Applications and Reviews,
  IEEE Transactions on}, 36(6):725--733, 2006.

\bibitem{chuy2007control}
O.~Y. Chuy, Y.~Hirata, Z.~Wang, and K.~Kosuge.
\newblock A control approach based on passive behavior to enhance user
  interaction.
\newblock {\em Robotics, IEEE Transactions on}, 23(5):899--908, 2007.

\bibitem{colombo2013motion}
A.~Colombo, D.~Fontanelli, A.~Legay, L.~Palopoli, and S.~Sedwards.
\newblock Motion planning in crowds using statistical model checking to enhance
  the social force model.
\newblock In {\em Decision and Control (CDC), 2013 IEEE 52nd Annual Conference
  on}, pages 3602--3608. IEEE, 2013.

\bibitem{ColomboFLPS13}
A.~Colombo, D.~Fontanelli, A.~Legay, L.~Palopoli, and S.~Sedwards.
\newblock {Motion Planning in Crowds using Statistical Model Checking to
  Enhance the Social Force Model}.
\newblock In {\em Proc. IEEE Int. Conf. on Decision and Control}, pages
  3602--3608, Florence, Italy, 10-13 Dec. 2013. IEEE.

\bibitem{ErVeJaDo2005}
J.~B. F.~V. Erp, H.~A. H. C.~V. Veen, C.~C.~Jansen, and T.~Dobbins.
\newblock Waypoint navigation with a vibrotactile waist belt.
\newblock {\em ACM Trans. Appl. Percept.}, 2(2):106--117, 2005.

\bibitem{FontanelliGPP13}
D.~Fontanelli, A.~Giannitrapani, L.~Palopoli, and D.~Prattichizzo.
\newblock {Unicycle Steering by Brakes: a Passive Guidance Support for an
  Assistive Cart}.
\newblock In {\em Proceedings of the 52nd IEEE International Conference on
  Decision and Control}, pages 2275--2280, 10-13 Dec. 2013.

\bibitem{gescheider94}
G.~A. Gescheider, S.~J. Bolanowski, K.~L. Hall, K.~E. Hoffman, and R.~T.
  Verrillo.
\newblock The effects of aging on information-processing channels in the sense
  of touch: I. absolute sensitivity.
\newblock {\em Somatosens Mot Res.}, 11(4):345--357, 1994.

\bibitem{ghosh2014following}
A.~Ghosh, L.~Alboul, J.~Penders, P.~Jones, and H.~Reed.
\newblock Following a robot using a haptic interface without visual feedback.
\newblock In {\em Proc. Int. Conf. on Advances in Computer-Human Interactions},
  pages 147--153, 2014.

\bibitem{Karuei-2011}
I.~Karuei, K.~E. MacLean, Z.~Foley-Fisher, R.~MacKenzie, S.~Koch, and
  M.~El-Zohairy.
\newblock Detecting vibrations across the body in mobile contexts.
\newblock In {\em Proc. Int. Conf. on Human Factors in Computing Systems},
  pages 3267--3276, 2011.

\bibitem{keates2003countering}
S.~Keates and J.~Clarkson.
\newblock Countering design exclusion.
\newblock In {\em Inclusive Design}, pages 438--453. Springer, 2003.

\bibitem{lee2010design}
G.~Lee, T.~Ohnuma, and N.~Y. Chong.
\newblock Design and control of jaist active robotic walker.
\newblock {\em Intelligent Service Robotics}, 3(3):125--135, 2010.

\bibitem{micaelli1993trajectory}
A.~Micaelli and C.~Samson.
\newblock Trajectory tracking for unicycle-type and two-steering-wheels mobile
  robots.
\newblock 1993.

\bibitem{NazemzadehFM13}
P.~Nazemzadeh, D.~Fontanelli, and D.~Macii.
\newblock {An Indoor Position Tracking Technique based on Data Fusion for
  Ambient Assisted Living}.
\newblock In {\em 2013 IEEE Intl. Conf. on Computational Intelligence and
  Virtual Environments for Measurement Systems and Applications}, pages 7--12,
  Milan, Italy, 15-17 July 2013. IEEE.

\bibitem{Naz14}
P.~Nazemzadeh, D.~Fontanelli, D.~Macii, and L.~Palopoli.
\newblock Indoor positioning of wheeled devices for ambient assisted living: A
  case study.
\newblock In {\em Instrumentation and Measurement Technology Conference (I2MTC)
  Proceedings, 2014 IEEE International}, pages 1421--1426, May 2014.

\bibitem{NazemzadehFMP14}
P.~Nazemzadeh, D.~Fontanelli, D.~Macii, and L.~Palopoli.
\newblock {Indoor Positioning of Wheeled Devices for Ambient Assisted Living: a
  Case Study}.
\newblock In {\em Proc. IEEE Int. Instrumentation and Measurement Technology
  Conference (I2MTC)}, pages 1421--1426, Montevideo, Uruguay, May 2014. IEEE.

\bibitem{pal15}
L.~Palopoli et~al.
\newblock Navigation assistance and guidance of older adults across complex
  public spaces: the {DALi} approach.
\newblock {\em Intelligent Service Robotics}, pages 1--16, 2015.

\bibitem{rizzon2013embedded}
L.~Rizzon and R.~Passerone.
\newblock Embedded soundscape rendering for the visually impaired.
\newblock In {\em Industrial Embedded Systems (SIES), 8th IEEE International
  Symposium on}, pages 101--104. IEEE, 2013.

\bibitem{rizzon2014spatial}
L.~Rizzon and R.~Passerone.
\newblock Spatial sound rendering for assisted living on an embedded platform.
\newblock In {\em Applications in Electronics Pervading Industry, Environment
  and Society}, volume 289 of {\em Lecture notes in Electrical Engineering},
  pages 61--73. Springer, 2014.

\bibitem{saida2011development}
M.~Saida, Y.~Hirata, and K.~Kosuge.
\newblock Development of passive type double wheel caster unit based on
  analysis of feasible braking force and moment set.
\newblock In {\em Proceedings of the 2011 IEEE/RSJ International Conference on
  Intelligent Robots and Systems}, pages 311--317, 2011.

\bibitem{ScAgPr-IMEKO14}
S.~Scheggi, M.~Aggravi, and D.~Prattichizzo.
\newblock A vibrotactile bracelet to improve the navigation of older adults in
  large and crowded environments.
\newblock In {\em Proc. 20th IMEKO TC4 Int. Symp. and 18th Int. Workshop on ADC
  Modelling and Testing Research on Electric and Electronic Measurement for the
  Economic Upturn}, pages 798--801, 2014.

\bibitem{ScMoPr-TH14}
S.~Scheggi, F.~Morbidi, and D.~Prattichizzo.
\newblock Human-robot formation control via visual and vibrotactile haptic
  feedback.
\newblock {\em IEEE Trans. on Haptics}, 2014.

\bibitem{SoetantoLP2003}
D.~Soetanto, L.~Lapierre, and A.~Pascoal.
\newblock Adaptive, non-singular path-following control of dynamic wheeled
  robots.
\newblock In {\em IEEE Conf. on Decision and Control}, volume~2, pages
  1765--1770. IEEE, 2003.

\bibitem{urdiales2011new}
C.~Urdiales, J.~M. Peula, M.~Fdez-Carmona, C.~Barru{\'e}, E.~J. P{\'e}rez,
  I.~S{\'a}nchez-Tato, J.~Del~Toro, F.~Galluppi, U.~Cort{\'e}s,
  R.~Annichiaricco, et~al.
\newblock A new multi-criteria optimization strategy for shared control in
  wheelchair assisted navigation.
\newblock {\em Autonomous Robots}, 30(2):179--197, 2011.

\bibitem{wasson2003user}
G.~Wasson, P.~Sheth, M.~Alwan, K.~Granata, A.~Ledoux, and C.~Huang.
\newblock User intent in a shared control framework for pedestrian mobility
  aids.
\newblock In {\em Intelligent Robots and Systems, 2003.(IROS 2003).
  Proceedings. 2003 IEEE/RSJ International Conference on}, volume~3, pages
  2962--2967. IEEE, 2003.

\bibitem{Weinstein68}
S.~Weinstein.
\newblock Intensive and extensive aspects of tactile sensitivity as a function
  of body part, sex, and laterality.
\newblock In {\em The skin senses}, pages 195--218. Erlbaum, 1968.

\bibitem{wilkinson2014applying}
C.~R. Wilkinson and A.~De~Angeli.
\newblock Applying user centred and participatory design approaches to
  commercial product development.
\newblock {\em Design Studies}, 35(6):614--631, 2014.

\bibitem{wilkinson2014demonstrating}
C.~R. Wilkinson, A.~De~Angeli, et~al.
\newblock Demonstrating a methodology for observing and documenting human
  behaviour and interaction.
\newblock In {\em DS 77: Proceedings of the DESIGN 2014 13th International
  Design Conference}, 2014.

\end{thebibliography}
% ACM needs 'a single self-contained file'!
%\balancecolumns

\balancecolumns
% That's all folks!
\end{document}